\documentclass[11pt]{article}

\PassOptionsToPackage{dvipsnames}{xcolor}
\PassOptionsToPackage{hyphens}{url}
\usepackage[preprint]{acl}

\usepackage{times}
\usepackage{latexsym}
\usepackage[T1]{fontenc}
\usepackage[utf8]{inputenc}
\usepackage{microtype}
\usepackage{inconsolata}
\usepackage{graphicx}

\usepackage{xspace}
\usepackage{amsmath}
\usepackage{amssymb}
\usepackage{dsfont}  
\usepackage{booktabs}
\usepackage[shortlabels,inline]{enumitem}

\usepackage{colortbl}
\usepackage{multirow}

\ifdefined\pdfminorversion
\fi

\ifdefined\pdfsuppressptexinfo
\fi

\DeclareMathOperator*{\argmax}{arg\,max}

\newcommand{\ours}{AdaptRouter\xspace}

\newcommand{\oursmethod}[1]{\cellcolor[HTML]{EAF3FF}\textbf{#1}}

\newcommand{\eqfontsize}[1]{\fontsize{#1}{\baselineskip}\selectfont}

\newsavebox{\equationbox}
\newcommand{\fitmath}[1]{%
  \begingroup
  \sbox{\equationbox}{$\displaystyle #1$}%
  \ifdim\wd\equationbox>\dimexpr\linewidth-2.5em\relax
    \resizebox{\dimexpr\linewidth-2.5em\relax}{!}{\usebox{\equationbox}}%
  \else
    \usebox{\equationbox}%
  \fi
  \endgroup
}

\usepackage[capitalize]{cleveref}

\makeatletter
\DeclareRobustCommand\onedot{\futurelet\@let@token\@onedot}
\def\@onedot{\ifx\@let@token.\else.\null\fi\xspace}

\makeatother

\title{AdaptRouter: Adaptive Routing across Evolving Tasks and Models}

\author{%
  \normalfont Jingpu Yang\textsuperscript{1,2}\quad
  Fengxian Ji\textsuperscript{3}\quad
  Tianhao Li\textsuperscript{3}\\
  \normalfont Juanfan Wu\textsuperscript{4}\quad
  Yufeng Wang\textsuperscript{1}\quad
  Shuxin Zheng\textsuperscript{2}\quad
  Xiaoqing Zhang\textsuperscript{2*}\\[0.6em]
  \normalfont\normalsize\textsuperscript{1}Beihang University\quad
  \textsuperscript{2}Beijing Zhongguancun Academy\\
  \normalfont\normalsize\textsuperscript{3}Northeastern University\quad
  \textsuperscript{4}Beijing Institute of Technology
}

\begin{document}
\maketitle
\begin{abstract}
Model routing assigns each query to a suitable foundation model, but tasks and available models keep changing, and existing routers are either untrained for the pool at hand or fit once and forget earlier tasks when updated.
We present \ours, an adaptive router trained by policy learning: a compact multimodal decision model reads each question with cards describing the current experts and learns from native task feedback, without routing labels, to select one frozen expert.
Open-weight and API experts share the router, its inputs, and the quality--cost trade-off but use different losses: cheap open-weight experts are explored with RL, whereas paid API experts are queried once and their cached feedback is reused as soft labels.
New tasks are learned without replay, new experts enter as cards, and an open-weight quality-first advantage provably keeps cost from overriding quality.
Across four task stages, \ours loses at most 0.4 points on earlier tasks, whereas feedback-trained routers lose up to 7.0, and as its pool grows from three to ten experts, its accuracy rises from 67.2\% to 72.1\%.
On five API models, it exceeds GPT-6.1 Sol and GPT-6 Astra (low) by over 22 points at under a third of their cost.
\end{abstract}

\section{Introduction}
\label{sec:intro}

\begin{figure}[t]
  \centering
  \includegraphics[width=\columnwidth]{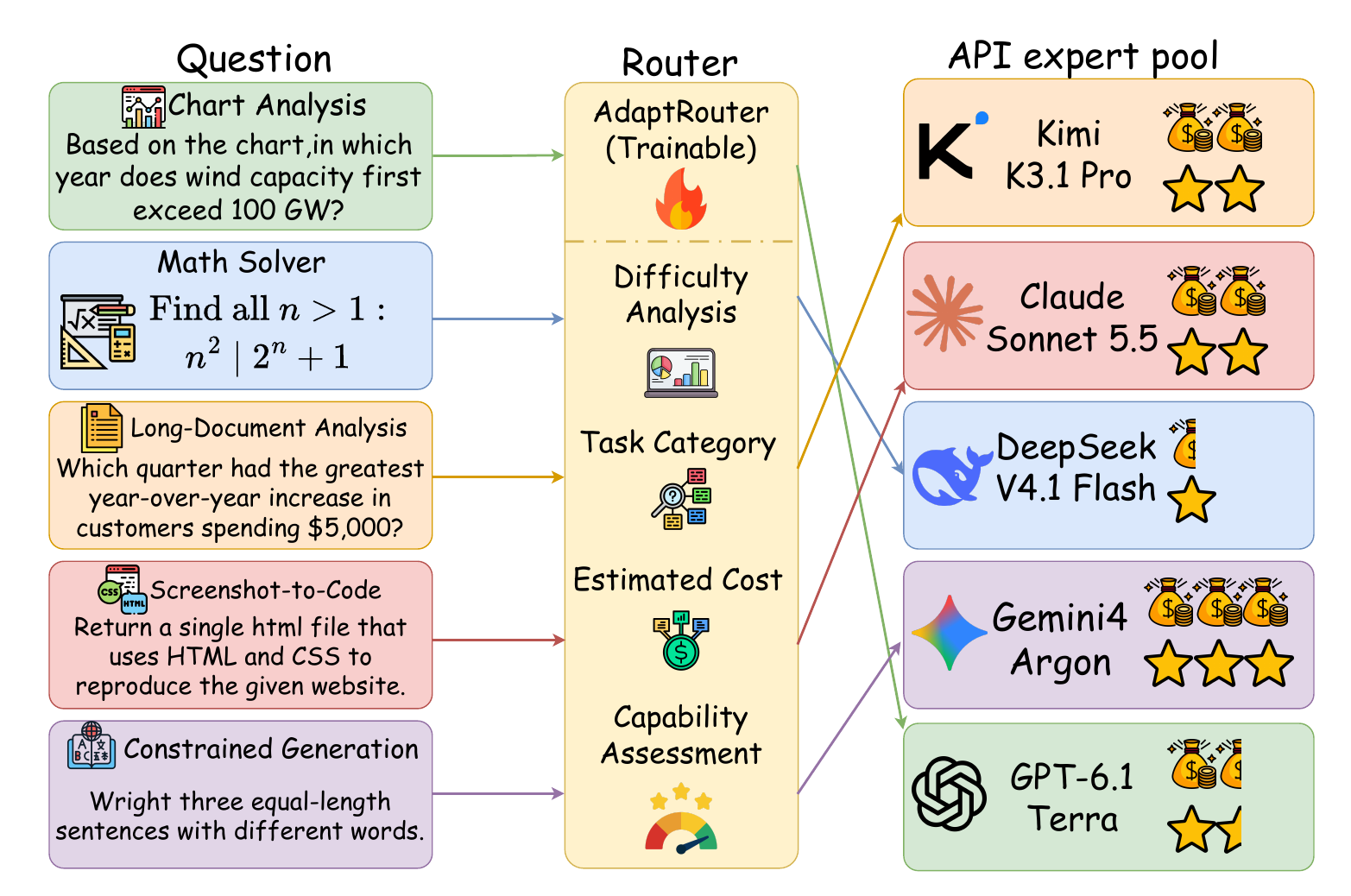}
  \caption{\textbf{Task-aware routing with \ours.}
  The router selects one expert per question from task requirements, expert capabilities, and cost.}
  \label{fig:teaser}
\end{figure}

\begin{figure*}[t]
  \centering
  \includegraphics[width=\textwidth]{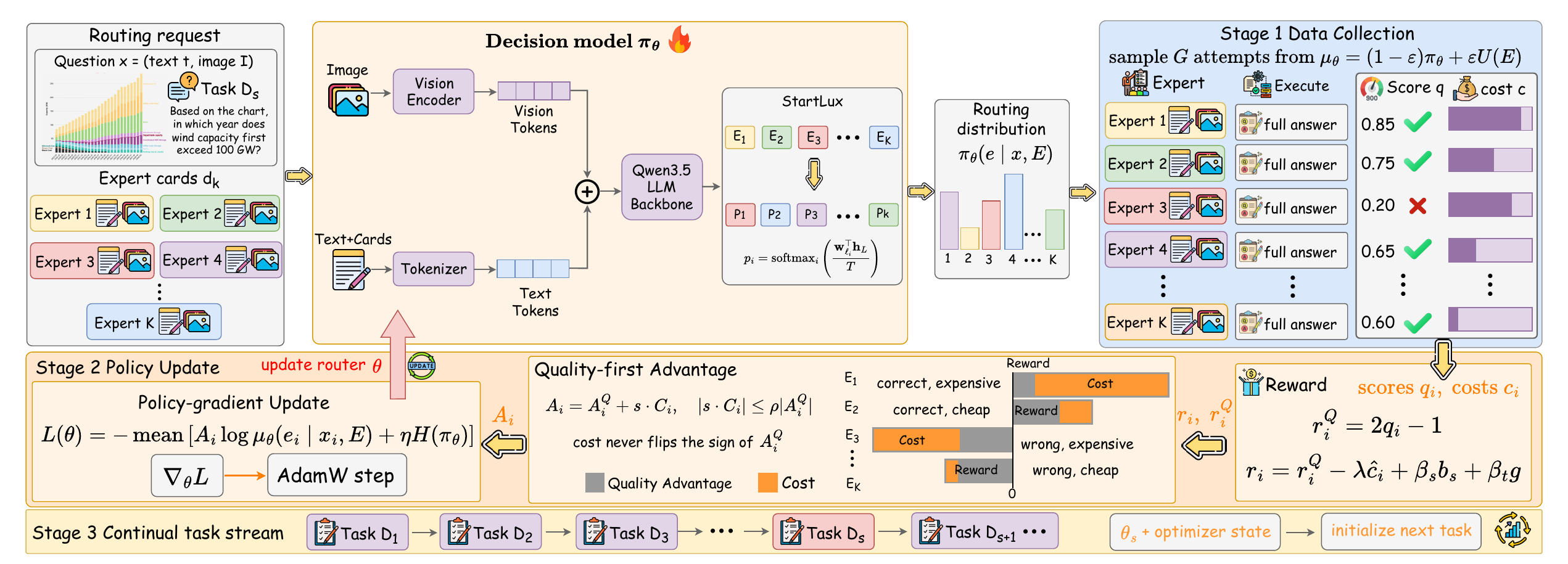}
  \caption{\textbf{Overview of \ours.}
  A decision model maps a question and expert cards to a policy $\pi_\theta$; in the open-weight setting, sampled experts are scored by quality and cost, and a quality-first advantage updates $\pi_\theta$.
  The API setting instead reuses collected expert feedback with soft-target cross-entropy and expected regret to avoid repeated paid exploration.
  Each adaptation stage inherits the previous policy and optimizer state; pool expansion updates the cards without changing the router architecture.}
  \label{fig:overview}
\end{figure*}

Foundation models and the tasks they serve keep changing: new models are released, existing ones are updated, repriced, or retired, and the questions that users bring shift over time.
Model routing assigns each question to a suitable model in a pool~\cite{lu2024zooter,li2026llmrouterbench,ma2026mmrbench} and exploits large differences in capability and price, which make the most expensive configuration often unnecessary.
On our open-weight suite, enabling reasoning in Qwen3.5-9B multiplies its parameter-weighted cost by about six but raises its average accuracy by only 1.5 points (\cref{tab:benchmark_results}); on our API suite, GPT-6 Astra at low reasoning effort costs five times as much as GPT-6.1 Sol for a gain of 0.3 points (\cref{tab:closed_source_results}).
Which expert suits which question, however, depends on the task mix and the pool, and both keep moving: a router fit once becomes stale, and refitting it on all past data after every change is costly, so the router itself has to adapt.

We therefore study adaptive routing: a router is updated as new tasks arrive and the expert pool changes, and it must stay accurate and economical throughout. This raises three requirements.
First, the router must learn each new task from limited feedback while preserving what it has learned, without revisiting earlier data.
Second, it must adapt to a changing pool: experts should enter as inputs rather than as fixed outputs, and a new pool should be learnable from a small exploration budget without routing labels.
Third, the router must predict from the input alone, including its images, which experts will answer correctly, and in open-weight exploration its learning signal must not let cost override correctness.

Existing routers meet these requirements only in part.
Released and prompted routers~\cite{tran2025archrouter,startlux,jevany,strands} select an expert from model descriptions or user-defined domains, so they accept a new pool, but they do not learn from its feedback.
Feedback-trained routers~\cite{ong2025routellm,zhuang2024embedllm,feng2025graphrouter,li2010linucb,zhou2026routeanything} learn to predict expert quality, mostly from frozen query embeddings with dedicated parameters per expert, and are fit once on a fixed split; updated on new tasks without replay, they forget earlier ones, losing up to 7.0 points in our experiments (\cref{tab:continual_results}).
Reinforcement-learning routers~\cite{zhang2025routerr1} train a language model to coordinate several calls on text-only tasks and fold cost into the reward as a weighted term, so a large cost weight can outweigh correctness.

We present \ours, an adaptive router trained by policy learning (\cref{fig:overview}): it feeds the complete question, including any images, together with short cards describing the current experts to a compact multimodal decision model (StartLux or a JevAny-style router, both 0.8B), which scores the listed candidates directly and selects one frozen expert to produce the entire answer (\cref{fig:teaser}).
In the open-weight setting, the policy learns from the native scores of its own sampled decisions, without routing labels, so the same exploration applies to any new task with an automatic scorer and to any new pool described by cards.
API experts, however, bill every call, and on-policy exploration would keep paying for the expert the policy already prefers while learning little about the alternatives.
We therefore query each API expert once per training question and reuse the cached feedback as reward-derived soft labels with a cross-entropy and expected-regret loss (\cref{sec:method_api}).
The two settings thus share the router, its inputs, and the quality--cost trade-off, but use different losses, chosen by the price of feedback.
Three components address the three requirements:
(i) stage-wise adaptation that carries parameters and optimizer state across stages, without replaying earlier datasets;
(ii) card-conditioned routing, in which a changed pool requires only new cards, not a new head; and
(iii) an open-weight quality-first learning signal that combines an engine-calibrated cost term with a leave-one-out group advantage in which cost ranks equally good experts but never reverses a quality preference.

\noindent Our contributions are:
\begin{itemize}
  \item \textbf{Adaptive routing framework.} \ours trains compact multimodal decision models as card-conditioned routers from native feedback, learning new tasks without replay and new experts without architectural change; its open-weight advantage provably keeps cost from overriding quality.
  \item \textbf{Adaptation to evolving tasks and pools.} \ours loses at most 0.4 points on earlier tasks (feedback-trained routers: up to 7.0), and its accuracy rises from 67.2\% to 72.1\% as its pool grows from three to ten experts; exploring only 5\% of the questions comes within 1.9 held-out points of budgets up to 90\%.
  \item \textbf{Accuracy at low cost.} On 15 text and vision benchmarks, \ours reaches 97\% of a 9B reasoning expert's accuracy at 12.6\% of its cost; on five API models, it exceeds GPT-6.1 Sol and GPT-6 Astra (low) by over 22 points at under a third of their cost.
\end{itemize}

\section{Related Work}
\label{sec:related}

\paragraph{Model routing.}
Early systems cascade models until an answer is accepted~\cite{chen2023frugalgpt} or choose between a small and a large model~\cite{ding2024hybridllm}.
Feedback-trained routers predict expert quality from reward models~\cite{lu2024zooter}, preference labels~\cite{ong2025routellm}, embeddings~\cite{zhuang2024embedllm}, graphs~\cite{feng2025graphrouter}, or bandit feedback~\cite{li2010linucb}; released and prompted routers select from domains or candidate lists without task-specific training~\cite{tran2025archrouter,startlux,jevany,strands}, and Router-R1~\cite{zhang2025routerr1} learns multi-call routing with RL.
Recent benchmarks and methods evaluate routing as quality minus priced cost~\cite{li2026llmrouterbench,ma2026mmrbench,dekoninck2025unified,zhou2026routeanything}.
\ours instead adapts a card-conditioned multimodal policy from native feedback, enforcing quality priority within open-weight exploration groups and reusing collected feedback for API training.

\paragraph{Adaptive routing and reinforcement learning.}
Adaptive routers keep past outcomes in external memory~\cite{zhou2026acrouter}, while feedback-trained routers are typically fit once~\cite{zhou2026routeanything}; updating a model on new data risks forgetting, which is commonly mitigated through regularization~\cite{kirkpatrick2017ewc,li2017lwf} or replay~\cite{rebuffi2017icarl,rolnick2019replay}, including for multimodal instruction tuning~\cite{he2023continual,chen2024coin}.
Reinforcement learning with verifiable rewards improves reasoning in language and vision--language models~\cite{shao2024deepseekmath,guo2025deepseekr1,liu2025visualrft,shen2025vlmr1} by optimizing generated answers.
In the open-weight setting, \ours optimizes one discrete decision per question with REINFORCE~\cite{williams1992reinforce} and a leave-one-out baseline~\cite{ahmadian2024rloo}, retaining earlier tasks without regularization, replay, or memory; the API setting uses a separate loss on collected feedback (\cref{sec:method_api}).

\paragraph{Efficient multimodal inference.}
Other methods reduce the cost of a single model by pruning visual tokens~\cite{chen2024fastv,shang2024prumerge,zhang2025sparsevlm,yang2025visionzip,wen2025tokenpruning}, sparse experts~\cite{shazeer2017moe,lin2024moellava}, dynamic depth~\cite{raposo2024mod,han2021dynamic}, or adaptive thinking~\cite{fang2025thinkless,zhang2025adaptthink}; \ours selects among unmodified models and is complementary to them.

\section{Problem Formulation}
\label{sec:problem}

A router is updated over stages $s=1,\dots,S$; at stage $s$, its task distribution $\mathcal{D}_s$, its pool $\mathcal{E}_s$ of frozen experts, or both may change as new tasks arrive or models are added, retired, or repriced.
Each expert $e_k$ is described by a short card $d_k$ (name, input modalities, reasoning mode; API cards also state token prices).

\paragraph{Routing objective.}
For a question $x=(t,\mathcal{I})$ with text $t$ and possibly empty images $\mathcal{I}$, the router $\pi_\theta(e\mid x,\mathcal{E}_s)$ commits to one expert, whose complete answer has native quality $q(x,e)\in[0,1]$ and cost $c(x,e)\ge0$.
At stage $s$, it maximizes
{\eqfontsize{9}
\begin{equation}
  \fitmath{%
  J_s(\theta)=\mathbb{E}_{x\sim\mathcal{D}_s}\,\mathbb{E}_{e\sim\pi_\theta(\cdot\mid x,\mathcal{E}_s)}\Big[\big(2q(x,e)-1\big)-\lambda\,\frac{c(x,e)}{\bar c}\Big],
  }
  \label{eq:objective}
\end{equation}
}%
where $\bar c$ is a reference cost (the mean cost of a reference expert, or of the training executions for API pools), so $\lambda$ trades $\lambda/2$ of accuracy for one unit of $\bar c$~\cite{dekoninck2025unified,zhou2026routeanything}.
No routing labels are given.

\paragraph{Adaptation protocol.}
\label{sec:problem_continual}
At stage $s$, the router inherits $\theta_{s-1}$ and its optimizer state. We study two sequences separately. In the task sequence (\cref{sec:exp_continual}), the pool is fixed, and each stage trains only on newly introduced datasets and is tested on all datasets seen so far. In the pool sequence (\cref{sec:exp_pool}), the pool grows from three to five to ten experts on a fixed set of training questions, every stage is tested on the same questions, and new experts enter only through their cards, without architectural change. We also adapt \ours to a separate pool of API models (\cref{sec:exp_api}).

\paragraph{Cost for comparison.}
Training uses the engine-calibrated cost of \cref{sec:method_reward}; for comparing open-weight experts, we report a parameter-weighted token cost~\cite{pope2023efficiently,srivatsa2024preble}
{\eqfontsize{9}
\begin{equation}
  \fitmath{%
  C=\textstyle\sum_{x} N(e_x)\bigl(n_{\mathrm{in}}(x,e_x)+n_{\mathrm{out}}(x,e_x)\bigr),
  }
  \label{eq:cost_proxy}
\end{equation}
}%
where $N(e_x)$ is the selected expert's size in billions and $n_{\mathrm{in}},n_{\mathrm{out}}$ count its input and generated tokens; API experts use token cost at list prices.

\begin{table*}[t]
  \centering
  \caption{Main-suite scores (\%). Avg.\ is request-weighted. Cost is relative to Qwen3.5-9B (thinking).
  \ours uses fully finetuned 0.8B routers (LoRA in \cref{tab:alternative_benchmark_results}).}
  \label{tab:benchmark_results}
  \setlength{\tabcolsep}{3pt}
  \renewcommand{\arraystretch}{1.16}
  \resizebox{\linewidth}{!}{%
  \begin{tabular}{@{}l*{18}{c}@{}}
    \toprule
    & \multicolumn{5}{c}{\cellcolor[HTML]{FFF2CC}LLMRouterBench} & \multicolumn{6}{c}{\cellcolor[HTML]{DCEBF7}MMR-Bench V2} & \multicolumn{4}{c}{\cellcolor[HTML]{E4EFD8}RobustRouteBench} & \cellcolor[HTML]{FCE8E6} & \cellcolor[HTML]{F1E7FA} & \cellcolor[HTML]{D9F2F0} \\
     \multirow{-2}{*}{\raisebox{0.5ex}{Method}} & \cellcolor[HTML]{FFF2CC}\shortstack{Acc. $\uparrow$\\MathBench} & \cellcolor[HTML]{FFF2CC}\shortstack{Acc. $\uparrow$\\KOR-Bench} & \cellcolor[HTML]{FFF2CC}\shortstack{Acc. $\uparrow$\\SimpleQA} & \cellcolor[HTML]{FFF2CC}\shortstack{Acc. $\uparrow$\\FinQA} & \cellcolor[HTML]{FFF2CC}\shortstack{Acc. $\uparrow$\\MELD} & \cellcolor[HTML]{DCEBF7}\shortstack{Acc. $\uparrow$\\RealWorldQA} & \cellcolor[HTML]{DCEBF7}\shortstack{Acc. $\uparrow$\\OCRBench} & \cellcolor[HTML]{DCEBF7}\shortstack{Acc. $\uparrow$\\CharXiv} & \cellcolor[HTML]{DCEBF7}\shortstack{Acc. $\uparrow$\\MathVerse} & \cellcolor[HTML]{DCEBF7}\shortstack{Acc. $\uparrow$\\BLINK} & \cellcolor[HTML]{DCEBF7}\shortstack{aAcc. $\uparrow$\\HallusionBench} & \cellcolor[HTML]{E4EFD8}\shortstack{Acc. $\uparrow$\\HellaSwag} & \cellcolor[HTML]{E4EFD8}\shortstack{HSR $\uparrow$\\FollowBench-zh} & \cellcolor[HTML]{E4EFD8}\shortstack{Acc. $\uparrow$\\C-Eval} & \cellcolor[HTML]{E4EFD8}\shortstack{Acc. $\uparrow$\\AI2D} & \cellcolor[HTML]{FCE8E6}\multirow{-2}{*}{Avg.} & \cellcolor[HTML]{F1E7FA}\multirow{-2}{*}{Cost} & \cellcolor[HTML]{D9F2F0}\multirow{-2}{*}{\shortstack{Mean routing\\time (s) $\downarrow$}} \\
    \midrule
    \multicolumn{19}{c}{\textit{I. Direct and random baselines}} \\
    \midrule
    Qwen3.5-9B (non-thinking) & 95.60 & 57.60 & 6.60 & 37.60 & 60.60 & 71.86 & 83.56 & 60.90 & 72.03 & 61.40 & 73.55 & 89.20 & 46.41 & 82.80 & 82.40 & 65.81 & 16.81\% & - \\
    Qwen3.5-9B (thinking) & 96.60 & 74.00 & 6.40 & 35.80 & 59.80 & 68.46 & 82.11 & 57.40 & 69.48 & 61.80 & 74.95 & 90.60 & 63.21 & 85.20 & 82.20 & 67.30 & 100.00\% & - \\
    Random & 88.80 & 46.00 & 3.00 & 28.80 & 48.80 & 24.08 & 31.89 & 24.00 & 28.74 & 25.60 & 32.46 & 75.60 & 41.54 & 74.20 & 33.20 & 38.39 & 9.44\% & - \\
    \midrule
    \multicolumn{19}{c}{\textit{II. Released and prompted routing methods}} \\
    \midrule
    StartLux 0.8B (Original) & 84.80 & 40.20 & 3.00 & 24.60 & 45.00 & 31.02 & 47.22 & 44.40 & 35.89 & 51.80 & 54.71 & 65.40 & 36.79 & 67.40 & 42.00 & 44.09 & 12.03\% & 0.267 \\
    JevAny-4B (Original) & 86.20 & 49.00 & 6.20 & 27.80 & 51.40 & 61.91 & 72.33 & 52.10 & 41.89 & 55.80 & 61.12 & 80.00 & 45.00 & 71.40 & 65.60 & 55.24 & 16.90\% & 0.314 \\
    Arch (Image-to-Text) & 81.00 & 27.00 & 0.40 & 5.20 & 0.00 & 3.01 & 22.67 & 3.10 & 1.79 & 3.40 & 9.42 & 21.00 & 34.36 & 67.60 & 7.60 & 17.92 & 5.42\% & 12.329 \\
    Qwen3.5-4B (Prompted Selector) & 90.00 & 50.60 & 3.40 & 28.80 & 42.00 & 57.72 & 58.89 & 32.60 & 39.21 & 41.20 & 52.10 & 78.80 & 44.23 & 71.20 & 43.60 & 48.31 & 8.64\% & 0.850 \\
    Strands Qwen3.5-2B (Decider) & 84.20 & 37.80 & 2.40 & 20.20 & 42.40 & 54.84 & 24.56 & 21.80 & 28.99 & 58.00 & 45.29 & 72.00 & 34.36 & 63.20 & 37.00 & 39.72 & 6.42\% & 0.173 \\
    \midrule
    \multicolumn{19}{c}{\textit{III. Feedback-trained routing methods}} \\
    \midrule
    Embedding-kNN & 93.80 & 53.80 & 4.00 & 27.20 & 59.80 & 21.99 & 79.89 & 58.40 & 73.69 & 23.20 & 72.55 & 89.40 & 48.08 & 75.40 & 85.00 & 57.91 & 11.86\% & 0.066 \\
    MLP Reward Predictor & 96.00 & 58.00 & 6.60 & 32.00 & 60.20 & 42.93 & 82.78 & 60.00 & 73.05 & 30.40 & 74.15 & 89.40 & 46.28 & 80.40 & 85.80 & 61.46 & 12.72\% & 0.064 \\
    EmbedLLM (adapted) & 95.40 & 58.00 & 6.00 & 34.60 & 60.00 & 43.98 & 83.22 & 60.00 & 73.44 & 42.60 & 73.75 & 89.40 & 46.92 & 79.20 & 86.20 & 62.33 & 12.74\% & 0.064 \\
    LinUCB & 96.20 & 53.00 & 5.80 & 31.00 & 60.20 & 57.72 & 82.78 & 60.10 & 74.84 & 40.20 & 73.35 & 88.80 & 45.51 & 80.80 & 84.20 & 62.81 & 12.62\% & 0.064 \\
    LatentRouter (adapted) & 95.80 & 57.00 & 4.00 & 23.80 & 60.20 & 10.21 & 83.00 & 60.60 & 72.54 & 1.80 & 74.75 & 89.00 & 47.05 & 81.20 & 83.80 & 56.60 & 12.65\% & 0.067 \\
    \midrule
    \multicolumn{19}{c}{\textit{IV. RL-trained routing methods}} \\
    \midrule
    \rowcolor[HTML]{EAF3FF}
    \oursmethod{\ours (StartLux)} & 96.20 & 58.40 & 5.80 & 37.60 & 60.20 & 69.24 & 83.00 & 57.70 & 72.67 & 61.20 & 72.55 & 89.20 & 46.41 & 81.00 & 82.00 & 65.08 & 12.56\% & 0.258 \\
    \rowcolor[HTML]{EAF3FF}
    \oursmethod{\ours (JevAny)} & 95.20 & 58.20 & 4.20 & 36.00 & 55.80 & 67.67 & 81.33 & 54.70 & 66.92 & 61.20 & 71.14 & 85.80 & 44.49 & 80.20 & 82.20 & 63.04 & 10.88\% & 0.208 \\
    \bottomrule
  \end{tabular}}
\end{table*}

\begin{table*}[t]
  \caption{Routing scores under task adaptation (\%). Stage-4 parentheses: change from each dataset's first stage; Cost and \ours rows as in \cref{tab:benchmark_results}; LoRA variants in \cref{tab:alternative_continual_results}. S4 mean routing time follows \cref{tab:benchmark_results} on the final-stage cohort.}
  \label{tab:continual_results}
  \centering
  \begingroup
  \definecolor{stagegain}{HTML}{009C9C}
  \definecolor{stageloss}{HTML}{E53935}
  \definecolor{stageneutral}{gray}{0.45}
  \newcommand{\stagechange}[2]{\,{\scriptsize\textcolor{#1}{(#2)}}}
  \footnotesize
  \setlength{\tabcolsep}{2pt}
  \renewcommand{\arraystretch}{1.09}
  \resizebox{\textwidth}{!}{%
  \begin{tabular}{@{}l*{16}{r}c@{}}
    \toprule
    & \multicolumn{2}{c}{\cellcolor[HTML]{FFF2CC}Stage 1} & \multicolumn{3}{c}{\cellcolor[HTML]{DCEBF7}Stage 2} & \multicolumn{4}{c}{\cellcolor[HTML]{E4EFD8}Stage 3} & \multicolumn{5}{c}{\cellcolor[HTML]{FCE8E6}Stage 4 (change vs.\ first)} & \multicolumn{1}{c}{\cellcolor[HTML]{F1E7FA}} & \multicolumn{1}{c}{\cellcolor[HTML]{FCE8E6}} & \multicolumn{1}{c}{\cellcolor[HTML]{D9F2F0}} \\
      \multirow{-2}{*}{\raisebox{0.5ex}{Method}} & \cellcolor[HTML]{FFF2CC}\shortstack{Acc. $\uparrow$\\MATH-\\500} & \cellcolor[HTML]{FFF2CC}\shortstack{Acc. $\uparrow$\\Chart\\QA} & \cellcolor[HTML]{DCEBF7}\shortstack{Acc. $\uparrow$\\MATH-\\500} & \cellcolor[HTML]{DCEBF7}\shortstack{Acc. $\uparrow$\\Chart\\QA} & \cellcolor[HTML]{DCEBF7}\shortstack{Acc. $\uparrow$\\MMLU-\\Pro} & \cellcolor[HTML]{E4EFD8}\shortstack{Acc. $\uparrow$\\MATH-\\500} & \cellcolor[HTML]{E4EFD8}\shortstack{Acc. $\uparrow$\\Chart\\QA} & \cellcolor[HTML]{E4EFD8}\shortstack{Acc. $\uparrow$\\MMLU-\\Pro} & \cellcolor[HTML]{E4EFD8}\shortstack{Acc. $\uparrow$\\Human\\Eval+} & \cellcolor[HTML]{FCE8E6}\shortstack{Acc. $\uparrow$\\MATH-\\500} & \cellcolor[HTML]{FCE8E6}\shortstack{Acc. $\uparrow$\\Chart\\QA} & \cellcolor[HTML]{FCE8E6}\shortstack{Acc. $\uparrow$\\MMLU-\\Pro} & \cellcolor[HTML]{FCE8E6}\shortstack{Acc. $\uparrow$\\Human\\Eval+} & \cellcolor[HTML]{FCE8E6}\shortstack{Acc. $\uparrow$\\IF\\Eval} & \multicolumn{1}{c}{\cellcolor[HTML]{F1E7FA}\multirow{-2}{*}{S4 cost}} & \multicolumn{1}{c}{\cellcolor[HTML]{FCE8E6}\multirow{-2}{*}{S4 Avg.}} & \multicolumn{1}{c}{\cellcolor[HTML]{D9F2F0}\multirow{-2}{*}{\shortstack{S4 mean routing\\time (s) $\downarrow$}}} \\
    \midrule
    \multicolumn{18}{c}{\textit{I. Direct and random baselines}} \\
    \midrule
    Qwen3.5-9B (non-thinking) & 92.60 & 85.88 & 92.60 & 85.88 & 74.40 & 92.60 & 85.88 & 74.40 & 84.15 & 92.60\stagechange{stageneutral}{+0.00} & 85.88\stagechange{stageneutral}{+0.00} & 74.40\stagechange{stageneutral}{+0.00} & 84.15\stagechange{stageneutral}{+0.00} & 85.40 & 20.15\% & 84.49 & - \\
    Qwen3.5-9B (thinking) & 73.40 & 86.08 & 73.40 & 86.08 & 77.60 & 73.40 & 86.08 & 77.60 & 89.02 & 73.40\stagechange{stageneutral}{+0.00} & 86.08\stagechange{stageneutral}{+0.00} & 77.60\stagechange{stageneutral}{+0.00} & 89.02\stagechange{stageneutral}{+0.00} & 91.87 & 100.00\% & 83.59 & - \\
    Random & 53.60 & 14.88 & 53.60 & 14.88 & 58.20 & 53.60 & 14.88 & 58.20 & 68.29 & 53.60\stagechange{stageneutral}{+0.00} & 14.88\stagechange{stageneutral}{+0.00} & 58.20\stagechange{stageneutral}{+0.00} & 68.29\stagechange{stageneutral}{+0.00} & 77.26 & 13.09\% & 54.45 & - \\
    \midrule
    \multicolumn{18}{c}{\textit{II. Released and prompted routing methods}} \\
    \midrule
    StartLux 0.8B (Original) & 39.80 & 21.44 & 39.80 & 21.44 & 42.40 & 39.80 & 21.44 & 42.40 & 61.59 & 39.80\stagechange{stageneutral}{+0.00} & 21.44\stagechange{stageneutral}{+0.00} & 42.40\stagechange{stageneutral}{+0.00} & 61.59\stagechange{stageneutral}{+0.00} & 75.60 & 20.63\% & 48.17 & 0.253 \\
    JevAny-4B (Original) & 42.20 & 31.00 & 42.20 & 31.00 & 51.80 & 42.20 & 31.00 & 51.80 & 73.78 & 42.20\stagechange{stageneutral}{+0.00} & 31.00\stagechange{stageneutral}{+0.00} & 51.80\stagechange{stageneutral}{+0.00} & 73.78\stagechange{stageneutral}{+0.00} & 84.10 & 24.97\% & 56.58 & 0.349 \\
    Arch (Image-to-Text) & 30.80 & 5.84 & 30.80 & 5.84 & 33.00 & 30.80 & 5.84 & 33.00 & 42.07 & 30.80\stagechange{stageneutral}{+0.00} & 5.84\stagechange{stageneutral}{+0.00} & 33.00\stagechange{stageneutral}{+0.00} & 42.07\stagechange{stageneutral}{+0.00} & 12.01 & 13.37\% & 24.74 & 9.061 \\
    Qwen3.5-4B (Prompted Selector) & 53.20 & 20.52 & 53.20 & 20.52 & 62.60 & 53.20 & 20.52 & 62.60 & 66.46 & 53.20\stagechange{stageneutral}{+0.00} & 20.52\stagechange{stageneutral}{+0.00} & 62.60\stagechange{stageneutral}{+0.00} & 66.46\stagechange{stageneutral}{+0.00} & 82.26 & 14.37\% & 57.01 & 0.801 \\
    Strands Qwen3.5-2B (Decider) & 59.60 & 12.60 & 59.60 & 12.60 & 47.00 & 59.60 & 12.60 & 47.00 & 56.10 & 59.60\stagechange{stageneutral}{+0.00} & 12.60\stagechange{stageneutral}{+0.00} & 47.00\stagechange{stageneutral}{+0.00} & 56.10\stagechange{stageneutral}{+0.00} & 72.27 & 9.28\% & 49.51 & 0.185 \\
    \midrule
    \multicolumn{18}{c}{\textit{III. Feedback-trained routing methods}} \\
    \midrule
    Embedding-kNN & 73.60 & 38.68 & 74.20 & 39.08 & 66.60 & 75.20 & 39.20 & 67.20 & 85.37 & 74.40\stagechange{stagegain}{+0.80} & 38.92\stagechange{stagegain}{+0.24} & 68.00\stagechange{stagegain}{+1.40} & 85.37\stagechange{stageneutral}{+0.00} & 84.47 & 12.71\% & 70.23 & 0.062 \\
    MLP Reward Predictor & 75.80 & 37.84 & 77.00 & 38.48 & 72.20 & 77.40 & 38.60 & 74.20 & 86.59 & 71.00\stagechange{stageloss}{-4.80} & 37.96\stagechange{stagegain}{+0.12} & 75.20\stagechange{stagegain}{+3.00} & 85.98\stagechange{stageloss}{-0.61} & 85.40 & 17.70\% & 71.11 & 0.060 \\
    EmbedLLM (adapted) & 74.80 & 38.28 & 75.60 & 38.48 & 75.60 & 76.40 & 38.56 & 75.60 & 88.41 & 77.20\stagechange{stagegain}{+2.40} & 38.48\stagechange{stagegain}{+0.20} & 75.00\stagechange{stageloss}{-0.60} & 89.63\stagechange{stagegain}{+1.22} & 82.81 & 15.71\% & 72.62 & 0.060 \\
    LinUCB & 76.20 & 38.80 & 74.00 & 38.08 & 67.80 & 73.60 & 37.96 & 67.80 & 84.15 & 71.80\stagechange{stageloss}{-4.40} & 38.12\stagechange{stageloss}{-0.68} & 68.80\stagechange{stagegain}{+1.00} & 82.93\stagechange{stageloss}{-1.22} & 82.26 & 9.37\% & 68.78 & 0.060 \\
    LatentRouter (adapted) & 64.00 & 37.68 & 63.40 & 37.64 & 43.80 & 63.40 & 36.96 & 43.80 & 51.83 & 57.00\stagechange{stageloss}{-7.00} & 37.56\stagechange{stageloss}{-0.12} & 75.20\stagechange{stagegain}{+31.40} & 77.44\stagechange{stagegain}{+25.61} & 85.21 & 18.84\% & 66.48 & 0.064 \\
    \midrule
    \multicolumn{18}{c}{\textit{IV. RL-trained routing methods}} \\
    \midrule
    \rowcolor[HTML]{EAF3FF}
    \oursmethod{\ours (StartLux)} & 92.20 & 85.84 & 91.80 & 85.44 & 74.00 & 92.00 & 85.52 & 73.80 & 84.10 & 92.20\stagechange{stageneutral}{+0.00} & 86.02\stagechange{stagegain}{+0.18} & 74.40\stagechange{stagegain}{+0.40} & 84.32\stagechange{stagegain}{+0.22} & 84.66 & 16.83\% & 84.32 & 0.250 \\
    \rowcolor[HTML]{EAF3FF}
    \oursmethod{\ours (JevAny)} & 83.20 & 83.58 & 90.20 & 84.91 & 71.80 & 91.00 & 84.63 & 72.80 & 80.49 & 92.00\stagechange{stagegain}{+8.80} & 85.84\stagechange{stagegain}{+2.26} & 74.40\stagechange{stagegain}{+2.60} & 81.71\stagechange{stagegain}{+1.22} & 83.92 & 16.73\% & 83.57 & 0.201 \\
    \bottomrule
  \end{tabular}}
  \endgroup
\end{table*}

\section{Method}
\label{sec:method}

In all settings, \ours routes with a policy $\pi_\theta$ of the same form: a compact multimodal decision model that reads the question and the expert cards, selects one frozen expert to answer, and is updated across stages (\cref{fig:overview}).
We study two ways of using feedback: for open-weight experts, the policy is updated from the quality and cost of its own sampled decisions (\cref{sec:method_reward,sec:method_optim}); for paid API experts, the answers and scores of all experts are collected in advance, and the policy is trained on this cached feedback (\cref{sec:method_api}).
Both settings share the routing inputs, the expert-selection interface, and the quality--cost trade-off of \cref{eq:objective}, but are trained with different losses.

\subsection{Decision Models as Card-conditioned Routers}
\label{sec:method_router}

\paragraph{Routing request.}
The router receives the question text, its original images (encoded by the router's own vision encoder), and expert cards, together with the objective stated in natural language. The main routing experiments randomize card order; the four-benchmark pool and exploration studies use a fixed registry order.
The cards are the only place where the pool enters the router: adding, retiring, or replacing an expert changes the cards without adding expert-specific parameters. Continued training updates the shared routing policy.

\paragraph{Policy.}
Our two 0.8B routers share a Qwen3.5 backbone~\cite{qwen35}: the released \emph{StartLux-0.8B}~\cite{startlux}, which scores candidates by the next-token logits of their letter labels, and a JevAny-style pointer router~\cite{jevany}. Both are fully finetuned in the main and task-adaptation experiments (LoRA~\cite{hu2022lora} variants in \cref{app:alternative_router}), whereas the pool and exploration studies train only the JevAny pointer head on a frozen backbone.
At test time, only $\hat e(x)=\argmax_{e}\pi_\theta(e\mid x,\mathcal{E})$ is executed, after a single prefill of the router.

\subsection{Open-weight Cost-aware Reward}
\label{sec:method_reward}

\paragraph{Engine-calibrated cost.}
Each execution's cost is estimated from its token counts with per-expert rates measured once on an inference engine~\cite{kwon2023vllm}:
{\eqfontsize{9}
\begin{equation}
  \fitmath{%
  c(x,e)=a_e+p_e n_{\mathrm{in}}+p'_e n_{\mathrm{in}}^2+v_e n_{\mathrm{vis}}
  +\sum_{k=1}^{n_{\mathrm{out}}-1}(o_e+o'_e(n_{\mathrm{in}}+k)),
  }
  \label{eq:cost}
\end{equation}
}%
covering request overhead, prefill, vision encoding, and decoding over a growing context; costs are normalized as $\hat c=c/\bar c$.

\paragraph{Reward.}
For an attempt that routes $x$ to $e$ and obtains native quality $q$, the reward is
{\eqfontsize{9}
\begin{equation}
  \fitmath{%
  r(x,e)=\underbrace{(2q-1)}_{r^{Q}}-\lambda\,\hat c(x,e)+\beta_{s}\,b_{s}(x,e)+\beta_{t}\,g(x,e).
  }
  \label{eq:reward}
\end{equation}
}%
The first two terms implement \cref{eq:objective}; the training-only shaping terms credit a correct answer from a small non-reasoning expert ($b_s$) and the margin by which the most expensive reasoning expert beats every cheaper expert ($g$), computed from all experts' scores on the training question.
We set $\lambda$ so that the cost spread of the pool is not negligible relative to the quality gap of $2$; otherwise the policy collapses to the most accurate expert.

\subsection{Open-weight Exploration and Policy Optimization}
\label{sec:method_optim}

For each training question, we draw $G$ independent attempts from the exploration policy $\mu_\theta=(1-\epsilon)\,\pi_\theta+\epsilon\,\mathcal{U}(\mathcal{E})$, execute each selected expert once, and score its complete answer.
Let $r_i$ and $r^{Q}_i$ denote the full reward and the quality term of attempt $i$ in a group.
Leave-one-out baselines~\cite{ahmadian2024rloo} yield the quality advantage and the full-reward advantage,
{\eqfontsize{9}
\begin{equation}
  \fitmath{%
  A^{Q}_i=r^{Q}_i-\tfrac{1}{G-1}\textstyle\sum_{j\neq i}r^{Q}_j,\qquad
  A^{R}_i=r_i-\tfrac{1}{G-1}\textstyle\sum_{j\neq i}r_j,
  }
\end{equation}
}%
and $C_i=A^{R}_i-A^{Q}_i$ is the component contributed by cost and shaping.
We combine them as $A_i=A^{Q}_i+s\,C_i$, where, if the qualities within the group differ,
{\eqfontsize{9}
\begin{equation}
  \fitmath{%
  s=\min\Big(1,\ \min_{i:\,C_i\neq 0}\frac{\rho\,|A^{Q}_i|}{|C_i|}\Big),\qquad \rho\in(0,1),
  }
  \label{eq:scale}
\end{equation}
}%
and $s=1$ if all qualities are equal or all $C_i=0$.
Groups in which no attempt obtains positive quality receive zero advantage.

This construction guarantees \emph{quality priority}: for every attempt $i$ with $A^{Q}_i\neq 0$, $\operatorname{sign}(A_i)=\operatorname{sign}(A^{Q}_i)$ and $|A_i-A^{Q}_i|\le\rho\,|A^{Q}_i|$.
To see this, note that if all qualities are equal, every $A^{Q}_i=0$ and the claim is vacuous.
Otherwise, \cref{eq:scale} gives $s|C_i|\le\rho|A^{Q}_i|$ for every $i$ with $C_i\neq0$; the inequality also holds when $C_i=0$.
Thus $|A_i-A^{Q}_i|=s|C_i|\le\rho|A^{Q}_i|<|A^{Q}_i|$ whenever $A^{Q}_i\neq0$, so $A_i$ has the same sign as $A^{Q}_i$.

Cost and shaping thus only rank attempts of equal quality, whereas the composite advantage $A^{R}_i$ ($s=1$) can reverse quality preferences.
The router minimizes the policy-gradient loss~\cite{williams1992reinforce}
{\eqfontsize{9}
\begin{equation}
  \fitmath{%
  \mathcal{L}(\theta)=-\frac{1}{|\mathcal{B}|}\sum_{i\in\mathcal{B}}\Big[A_i\log\mu_\theta(e_i\mid x_i,\mathcal{E})+\eta\,\mathcal{H}\big(\pi_\theta(\cdot\mid x_i,\mathcal{E})\big)\Big]
  }
  \label{eq:loss}
\end{equation}
}%
with one AdamW step per batch; the entropy bonus keeps the policy from saturating on a single expert.
Because cost never reverses the sign of a quality advantage, every adaptation stage stays anchored to correctness.

\subsection{API Training with Reward-derived Soft Labels}
\label{sec:method_api}

On-policy exploration over paid API experts would repeatedly query, across training steps, the expert that the current policy already prefers on a question, incurring repeated charges while rarely observing feedback from the other experts.
We therefore query every candidate expert once on each training question and cache its answer, quality score, and list-price cost; subsequent optimization reads only these cached records and issues no further API calls.
With quality $q_k$, cost $c_k$, and mean training cost $\bar c$, the utility $u_k=q_k-\lambda c_k/\bar c$ of each expert $e_k$ follows the quality--cost form of \cref{eq:objective} and defines a soft target $y=\operatorname{softmax}(u/\tau)$ over all experts; with $p_k=\pi_\theta(e_k\mid x,\mathcal E)$, the router minimizes the soft-target cross-entropy plus the expected regret:
{\eqfontsize{9}
\begin{equation}
  \fitmath{%
  \mathcal L_{\mathrm{API}}(x;\theta)=-\sum_k y_k\log p_k
  +0.1\Big(\max_k u_k-\sum_k p_k u_k\Big).
  }
  \label{eq:api_loss}
\end{equation}
}%

Throughout, \emph{without routing labels} means that no per-question human annotation of the best expert is required; training still uses the quality and cost feedback on the experts' answers.
Unlike \cref{eq:loss}, \cref{eq:api_loss} has no shaping terms, so quality and cost trade off through $\lambda$ alone (\cref{app:api_training}), and the quality-priority guarantee of \cref{sec:method_optim} does not apply to it.

\section{Experiments}
\label{sec:experiments}

\begin{table*}[t]
  \centering
  \caption{Closed-source routing on 10,363 test questions. Avg.\ is weighted by test-set size.
  Cost is total expert inference cost at list prices (USD). Routing time is per question.}
  \label{tab:closed_source_results}
  \setlength{\tabcolsep}{3pt}
  \renewcommand{\arraystretch}{1.12}
  \resizebox{\textwidth}{!}{%
  \begin{tabular}{@{}l*{13}{c}@{}}
    \toprule
    \multirow{2}{*}{Method} & \multicolumn{10}{c}{\cellcolor[HTML]{E4EFD8}Benchmark scores (\%) $\uparrow$} & \cellcolor[HTML]{FCE8E6} & \cellcolor[HTML]{F1E7FA} & \cellcolor[HTML]{D9F2F0} \\
    & \cellcolor[HTML]{E4EFD8}\shortstack{Omni-\\MATH} & \cellcolor[HTML]{E4EFD8}\shortstack{BBEH\\Mini} & \cellcolor[HTML]{E4EFD8}\shortstack{HLE\\Text} & \cellcolor[HTML]{E4EFD8}IFBench & \cellcolor[HTML]{E4EFD8}\shortstack{AA-LCR\\v1.1} & \cellcolor[HTML]{E4EFD8}\shortstack{LongBench\\v2} & \cellcolor[HTML]{E4EFD8}\shortstack{MMMU-\\Pro} & \cellcolor[HTML]{E4EFD8}ChartQAPro & \cellcolor[HTML]{E4EFD8}\shortstack{Design2Code\\Hard} & \cellcolor[HTML]{E4EFD8}\shortstack{LiveCodeBench\\Pro} & \cellcolor[HTML]{FCE8E6}\multirow{-2}{*}{\shortstack{Weighted\\Avg. (\%) $\uparrow$}} & \cellcolor[HTML]{F1E7FA}\multirow{-2}{*}{\shortstack{Total cost $\downarrow$\\(USD)}} & \cellcolor[HTML]{D9F2F0}\multirow{-2}{*}{\shortstack{Mean routing\\time (s) $\downarrow$}} \\
    \midrule
    \multicolumn{14}{c}{\textit{I. Direct models}} \\
    \midrule
    GPT-6.1 Sol (low) & 64.32 & 20.49 & 17.77 & 48.00 & 66.00 & 60.18 & 62.39 & 26.79 & 82.36 & 69.15 & 47.23 & 429.79 & - \\
    GPT-6 Astra (low) & 65.52 & 21.71 & 16.99 & 44.40 & 74.00 & 57.30 & 63.51 & 26.97 & 84.05 & 62.77 & 47.57 & 2157.90 & - \\
    Gemini 4 Argon & 84.14 & 22.68 & 46.70 & 58.00 & 72.00 & 66.59 & 85.94 & 52.18 & 70.71 & 71.28 & 68.35 & 260.53 & - \\
    Claude Sonnet 5.5 & 84.89 & 30.00 & 33.01 & 38.80 & 62.00 & 63.50 & 77.31 & 23.01 & 73.42 & 55.32 & 59.74 & 257.26 & - \\
    GPT-6.1 Sol (max) & - & - & 52.92 & - & 83.00 & - & 85.95 & - & - & - & - & - & - \\
    GPT-6 Astra (max) & - & - & 54.68 & - & 80.67 & - & 86.88 & - & - & - & - & - & - \\
    \midrule
    \multicolumn{14}{c}{\textit{II. Released and prompted routing methods}} \\
    \midrule
    StartLux 0.8B (Original) & 79.30 & 21.71 & 34.55 & 51.20 & 74.00 & 56.42 & 76.85 & 51.65 & 77.56 & 68.09 & 62.14 & 196.70 & 0.257 \\
    JevAny-4B (Original) & 79.80 & 30.49 & 20.91 & 43.60 & 64.00 & 61.73 & 64.03 & 26.84 & 76.64 & 58.51 & 54.26 & 201.46 & 0.291 \\
    Qwen3.5-4B (Prompted Selector) & 76.46 & 40.73 & 30.23 & 65.20 & 76.00 & 56.86 & 68.80 & 45.36 & 76.31 & 69.15 & 59.19 & 75.01 & 0.485 \\
    \midrule
    \multicolumn{14}{c}{\textit{III. Feedback-trained routing methods}} \\
    \midrule
    MLP Reward Predictor & 86.27 & 19.76 & 46.55 & 58.00 & 62.00 & 58.41 & 84.76 & 48.34 & 82.91 & 73.40 & 67.90 & 246.87 & 0.053 \\
    LatentRouter (adapted) & 80.93 & 44.88 & 38.11 & 67.60 & 58.00 & 60.40 & 77.83 & 49.61 & 83.16 & 67.02 & 64.67 & 123.31 & 0.056 \\
    Router-R1 (PPO) & 86.78 & 23.66 & 48.92 & 60.00 & 72.00 & 66.81 & 88.29 & 53.80 & 79.70 & 74.47 & 70.53 & 283.26 & 0.315 \\
    \midrule
    \multicolumn{14}{c}{\textit{IV. Soft-label-trained routing methods}} \\
    \midrule
    \rowcolor[HTML]{EAF3FF}
    \oursmethod{\ours (StartLux)} & 86.68 & 49.02 & 48.66 & 66.80 & 82.00 & 62.83 & 87.44 & 53.42 & 83.43 & 74.47 & 71.31 & 120.46 & 0.283 \\
    \rowcolor[HTML]{EAF3FF}
    \oursmethod{\ours (JevAny)} & 86.60 & 33.90 & 48.92 & 64.40 & 82.00 & 62.83 & 88.36 & 53.75 & 83.43 & 74.47 & 70.86 & 134.30 & 0.217 \\
    \bottomrule
  \end{tabular}}
\end{table*}

We evaluate the three requirements of adaptive routing in turn: learning new tasks while retaining earlier ones (\cref{sec:exp_continual}), adapting to new expert pools (\cref{sec:exp_pool}), and accurate, low-cost routing on the main suite and on a new pool of API models (\cref{sec:exp_main,sec:exp_api}).

\subsection{Setup}
\label{sec:exp_setup}

\paragraph{Expert pools.}
NT and T denote non-reasoning and reasoning modes (outputs capped at 4,096/16,384 tokens).
M3 contains Qwen3.5-2B (NT), 4B (NT), and 9B (T)~\cite{qwen35}; M5 adds 4B (T) and 9B (NT); M10 adds Gemma-4-12B, MiniCPM-V-4.5 (NT), GLM-4.6V-Flash, Step3-VL-10B, and Ovis2.5-9B (T).
The main suite, the task sequence, and the ablations use M5.
The API pool contains GPT-6.1 Terra, DeepSeek V4.1 Flash, Claude Sonnet 5.5, Gemini 4 Argon, and Kimi K3.1 Pro; GPT-6.1 Sol and GPT-6 Astra serve as references.

\paragraph{Benchmarks.}
The task sequence has four stages: MATH-500 with ChartQA, then MMLU-Pro, HumanEval+, and IFEval~\cite{lightman2023verify,masry2022chartqa,wang2024mmlupro,liu2023evalplus,zhou2023ifeval}.
The pool, exploration, router-scale, and $\lambda$ studies use the 2,680 questions of MathBench, KOR-Bench, OCRBench, and FollowBench-zh; unless noted, 400 are used for training and the other 2,280 for testing.
The main suite has 15 text and vision benchmarks from LLMRouterBench, MMR-Bench, and RobustRouteBench~\cite{li2026llmrouterbench,ma2026mmrbench,robustroutebench}.
The API suite has ten: Omni-MATH~\cite{gao2024omnimath}, BBEH Mini~\cite{kazemi2025bbeh}, HLE (text)~\cite{phan2025hle}, IFBench~\cite{pyatkin2025ifbench}, AA-LCR v1.1~\cite{artificialanalysis2025aalcr}, LongBench v2~\cite{bai2025longbenchv2}, MMMU-Pro~\cite{yue2025mmmupro}, ChartQAPro~\cite{masry2025chartqapro}, Design2Code-Hard~\cite{si2025design2code}, and LiveCodeBench-Pro~\cite{zheng2025livecodebenchpro}, the last scored by GPT-6 Luna code review.

\paragraph{Baselines and metrics.}
We compare fixed experts, random routing, released or prompted selectors~\cite{startlux,jevany,tran2025archrouter,strands}, feedback-trained routers~\cite{zhuang2024embedllm,li2010linucb}, and, for the API pool, Router-R1~\cite{zhang2025routerr1} adapted to a single decision and trained with PPO from the same StartLux-0.8B on the same questions.
\ours (StartLux) and \ours (JevAny) are the routers of \cref{sec:method_router}.
Avg. is request-weighted; open-weight Cost follows \cref{eq:cost_proxy}, and API Cost is token cost at list prices.

\paragraph{Routing time.}
Mean routing time measures wall-clock expert selection time per question. In \cref{tab:benchmark_results}, baseline timings use request-weighted stratified samples. Arch timing includes image transcription, and feedback-trained routers use stage-4 checkpoints.
In \cref{tab:closed_source_results}, groups III and IV use a 388-question stratified sample weighted by benchmark size. Group III uses the MLP and LatentRouter validation checkpoints and the Router-R1 StartLux initialization. Group IV uses the retained global checkpoints.

\subsection{Adaptation to New Tasks}
\label{sec:exp_continual}

Learned routers are updated stage by stage on new datasets only, and \cref{tab:continual_results} reports their scores on all datasets seen so far.

\paragraph{Retention and adaptation.}
Relative to its best earlier score on each dataset, neither instantiation of \ours drops by more than 0.4 points at any stage, whereas at the final stage LinUCB, the MLP predictor, and the latent-factor router drop by 4.4, 6.4, and 7.0 points on MATH-500.
Each new dataset is learned in its stage: \ours lifts the released StartLux-0.8B from 39.80 to 92.20 on MATH-500 and from 21.44 to 85.84 on ChartQA.
After the final stage, the two routers (84.3 and 83.6 on average) match the fixed 9B experts (84.5 and 83.6) at 16.7--16.8\% of the reasoning expert's cost; LoRA variants behave similarly (\cref{tab:alternative_continual_results}).

\subsection{Adaptation to Expanding Expert Pools}
\label{sec:exp_pool}

A single JevAny-0.8B router (the JevAny-style router of \cref{sec:method_router}) is updated as its pool expands through M3 $\rightarrow$ M5 $\rightarrow$ M10. Each stage starts from the previous checkpoint, including its pointer head and AdamW state, trains on the same 400 questions (100 per benchmark), and is evaluated on the same 2,280 test questions (\cref{app:hyperparameters}). Only the pointer head is trained, and new experts enter only as cards, without expert-specific parameters.

\begin{table}[t]
  \centering
  \caption{Sequential pool expansion M3 $\rightarrow$ M5 $\rightarrow$ M10 (\%). NT/T: non-thinking/thinking.}
  \label{tab:pool}
  \small
  \setlength{\tabcolsep}{2pt}
  \begin{tabular}{@{}lccccc@{}}
    \toprule
    Method & \shortstack{Math-\\Bench} & \shortstack{KOR-\\Bench} & \shortstack{OCR-\\Bench} & \shortstack{Follow-\\Bench-zh} & All \\
    \midrule
    Qwen3.5-9B (NT) & 95.3 & 58.0 & 83.1 & 44.3 & 69.3 \\
    Qwen3.5-9B (T) & 96.3 & 76.5 & 81.4 & 61.8 & 77.3 \\
    Step3-VL-10B (T) & 96.5 & 80.0 & 79.9 & 40.7 & 71.1 \\
    Gemma-4-12B (NT) & 95.0 & 71.3 & 72.4 & 58.5 & 72.0 \\
    \midrule
    JevAny-0.8B (M3) & 93.0 & 67.0 & 78.9 & 38.4 & 67.2 \\
    JevAny-0.8B (M5) & 93.0 & 70.3 & 79.0 & 36.6 & 67.3 \\
    JevAny-0.8B (M10) & 92.3 & 74.8 & 84.5 & 44.3 & 72.1 \\
    \bottomrule
  \end{tabular}
\end{table}

\begin{figure}[t]
  \centering
  \includegraphics[width=\columnwidth]{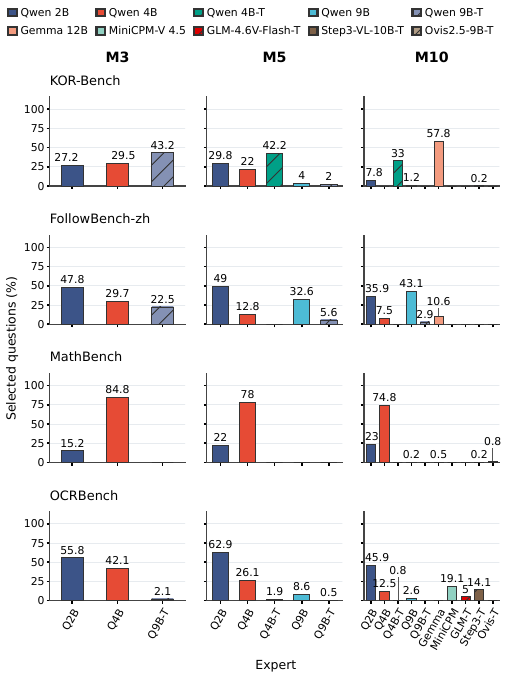}
  \caption{\textbf{Expert selection shares during pool expansion.}
  M5 continues from M3, and M10 from M5. Q denotes Qwen; T and hatching indicate reasoning mode.}
  \label{fig:pool}
\end{figure}

\paragraph{Using added experts.}
As the pool grows, accuracy rises from 67.2\% (M3) to 67.3\% (M5) and 72.1\% (M10) (\cref{tab:pool}), as the router shifts questions to the experts added at each stage (\cref{fig:pool}).
At M5, it sends 169 of 400 KOR-Bench questions to the new Qwen3.5-4B (T), and KOR-Bench rises from 67.0 to 70.3.
At M10, the five new experts take 27.0\% of all questions: Gemma-4-12B answers 231 KOR-Bench questions (74.8), and MiniCPM-V-4.5, GLM-4.6V-Flash, and Step3-VL-10B answer 306 OCRBench questions, where accuracy reaches 84.5, above every single expert in \cref{tab:pool}.
The same router thus adapts to newly added experts and turns a larger pool into higher accuracy.
With cumulative replay, All changes from 67.3\% to 70.0\% at M5 and from 72.1\% to 71.2\% at M10 (\cref{tab:pool_training_protocols}).

\begin{table}[t]
  \centering
  \caption{M10 accuracy (\%) by share of the 2,680 questions explored. All: the 2,280 test questions of \cref{tab:pool}. Held-out: fixed 10\%, never explored. Remain Unseen: all questions that run never explored. Non-Qwen: selection share.}
  \label{tab:data_scale}
  \small
  \setlength{\tabcolsep}{1.5pt}
  \resizebox{\linewidth}{!}{%
  \begin{tabular}{@{}l*{8}{c}@{}}
    \toprule
    Explored & \shortstack{Math-\\Bench} & \shortstack{KOR-\\Bench} & \shortstack{OCR-\\Bench} & \shortstack{Follow-\\Bench-zh} & All & \shortstack{Held-out\\(10\%)} & \shortstack{Remain\\Unseen} & \shortstack{Non-\\Qwen} \\
    \midrule
    1\% & 92.3 & 67.0 & 84.4 & 38.7 & 69.1 & 73.1 & 70.2 & 27.5 \\
    5\% & 91.5 & 76.0 & 81.8 & 47.1 & 72.1 & 77.2 & 72.9 & 33.7 \\
    10\% & 93.5 & 76.8 & 84.4 & 53.8 & 75.5 & 77.2 & 75.4 & 35.8 \\
    15\% & 93.8 & 72.3 & 85.9 & 54.6 & 75.5 & 79.1 & 75.5 & 43.6 \\
    20\% & 93.3 & 76.0 & 85.6 & 56.8 & 76.7 & 78.7 & 76.6 & 50.7 \\
    50\% & 93.5 & 76.5 & 86.6 & 59.7 & 78.0 & 78.0 & 76.9 & 62.5 \\
    60\% & 93.8 & 77.8 & 86.3 & 58.1 & 77.7 & 78.7 & 77.2 & 48.0 \\
    70\% & 93.3 & 76.8 & 86.5 & 59.0 & 77.8 & 78.4 & 78.4 & 56.8 \\
    80\% & 93.8 & 76.8 & 87.9 & 59.0 & 78.3 & 77.6 & 80.8 & 73.8 \\
    90\% & 93.3 & 77.3 & 87.0 & 58.5 & 77.9 & 77.6 & 77.6 & 58.6 \\
    \bottomrule
  \end{tabular}}
\end{table}

\paragraph{Exploration budget for a new pool.}
\Cref{tab:data_scale} trains independent M10 routers (500 updates each) that explore 1--90\% of the questions and learn only from their own decisions.
To separate adaptation after exploration from generalization to unexplored questions, we report three evaluation sets.
All, the 2,280 test questions of \cref{tab:pool}, also contains explored questions at budgets of 20\% and above, and thus describes overall performance on a mixed set.
Held-out, a fixed set of question families covering 10\% of the questions, is identical across budgets and never explored; it is our primary basis for comparing generalization across budgets.
Remain Unseen contains all questions left unexplored at a given budget, so its composition changes with the budget.
Held-out accuracy reaches 77.2\% at 5\% exploration and stays within 77.2--79.1\% up to 90\%, so a small budget suffices to learn a new pool.
At 20--90\%, where All includes explored questions, accuracy on the never-explored questions (Remain Unseen) stays within $-1.1$ to $+2.5$ points of All, so All is not inflated by exploration.

\subsection{Routing Quality and Cost}
\label{sec:exp_main}

On the main suite (\cref{tab:benchmark_results}), \ours lifts StartLux-0.8B from 44.09 to 65.08; \ours (JevAny) reaches 63.04, above JevAny-4B (55.24).
StartLux reaches 96.7\% of the 9B reasoning expert's accuracy at 12.56\% of its cost and is within 0.73 points of the 9B non-reasoning expert at three quarters of its cost.
Feedback-trained routers reach at most 62.81, below both \ours routers, at 11.86--12.74\% cost; \ours (Jev\-Any) is cheaper than all of them (10.88\%).

\begin{figure}[t]
  \centering
  \includegraphics[width=\columnwidth]{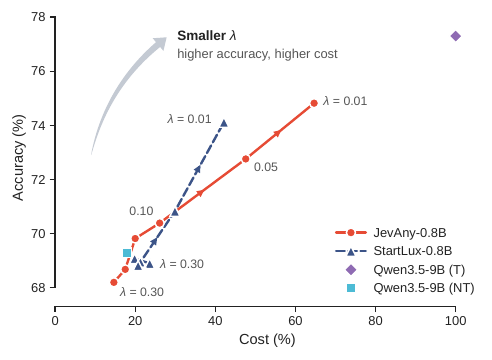}
  \caption{\textbf{Effect of the trade-off coefficient $\lambda$.}
  Accuracy versus cost of routers trained with different $\lambda$ on the M5 expert pool (four-benchmark test questions).
  Cost is shown as a percentage of the Qwen3.5-9B (T) cost.}
  \label{fig:lambda}
\end{figure}

\begin{figure}[t]
  \centering
  \includegraphics[width=\columnwidth]{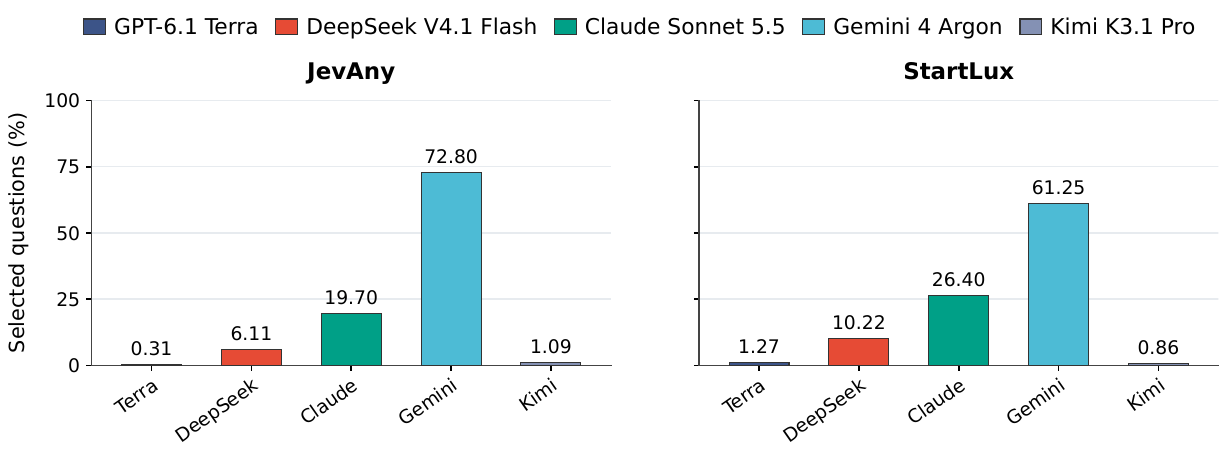}
  \caption{API model selection shares of JevAny (left) and StartLux (right) in \cref{tab:closed_source_results}.}
  \label{fig:api_selection_share}
\end{figure}

\subsection{Adapting to a Pool of API Models}
\label{sec:exp_api}

\paragraph{Training on the API pool.}
This experiment examines whether the same question--card routing framework carries over to paid API experts.
Following \cref{sec:method_api}, both routers are trained with \cref{eq:api_loss} on feedback collected once from a new pool of five API models (none priced above GPT-6.1 Sol per token) on 1,328 training questions, and optimization issues no further expert API calls.
\Cref{tab:closed_source_results} reports their answer quality and inference cost on 10,363 test questions from the ten benchmarks, drawn from question groups disjoint from the training questions; long inputs reach the router as the question, its length, and document excerpts (\cref{app:api_training}).

\paragraph{Quality and cost.}
StartLux reaches the highest weighted average, 71.31 (JevAny: 70.86), 24.08 and 23.74 points above GPT-6.1 Sol and GPT-6 Astra (low) at 28\% and 6\% of their cost.
It also beats Gemini 4 Argon (68.35) and Claude Sonnet 5.5 (59.74) used alone on nine benchmarks each at less than half of their cost, and Router-R1, trained from the same StartLux-0.8B with PPO, reaches 70.53 at 2.35 times StartLux's cost.
Together with \cref{sec:exp_main}, these results indicate that the card-conditioned design is effective under both training losses and does not depend on RL.

\paragraph{Where the savings come from.}
StartLux sends nearly nine in ten questions to Gemini 4 Argon or Claude Sonnet 5.5 (\cref{fig:api_selection_share}) yet costs less than half as much as either, because cost concentrates on long contexts: LongBench v2 (4.36\% of questions) costs \$112.28 with Gemini 4 Argon, whereas StartLux sends almost all of it to DeepSeek V4.1 Flash for \$9.54.

\subsection{Analysis}
\label{sec:exp_analysis}

\begin{figure}[t]
  \centering
  \includegraphics[width=\columnwidth]{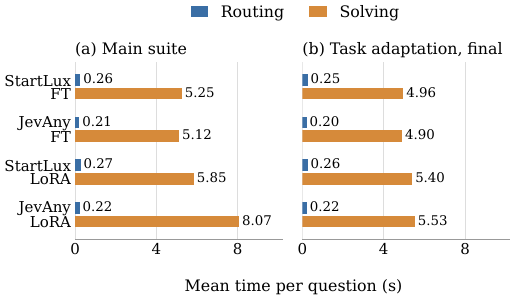}
  \caption{\textbf{Routing and expert execution time.}
  Mean time per question on the main suite (\cref{tab:benchmark_results}) and the final task-adaptation stage (\cref{tab:continual_results}).}
  \label{fig:routing_solving_time}
\end{figure}

\paragraph{Routing time and $\lambda$.}
Mean routing takes 0.20--0.27 seconds per question, while the selected experts take 4.90--8.07 GPU seconds (\cref{fig:routing_solving_time}).
As $\lambda$ decreases from 0.30 to 0.01, StartLux rises from 68.9\% to 74.1\% accuracy while its cost grows from 23.6\% to 42.2\% of Qwen3.5-9B (T) (\cref{fig:lambda}); details are in \cref{app:tradeoff_details}.

\begin{table}[t]
  \centering
  \caption{Router scale on M5 (\%).
  Mean/p95: router latency per question incl.\ preprocessing. Order consist.: pairwise agreement of selected experts across five cyclic candidate orders, measuring the stability of the selection.}
  \label{tab:router_scale}
  \small
  \setlength{\tabcolsep}{2pt}
  \renewcommand{\arraystretch}{1.12}
  \resizebox{\linewidth}{!}{%
  \begin{tabular}{@{}l*{8}{c}@{}}
    \toprule
    Size & \shortstack{Math-\\Bench} & \shortstack{KOR-\\Bench} & \shortstack{OCR-\\Bench} & \shortstack{Follow-\\Bench-zh} & All & \shortstack{Mean\\(ms)} & \shortstack{p95\\(ms)} & \shortstack{Order\\consist.\\(\%)} \\
    \midrule
    \multicolumn{9}{c}{\textit{StartLux (LoRA)}} \\
    \midrule
    0.8B & 93.00 & 66.75 & 80.25 & 42.79 & 68.95 & 262.6 & 343.1 & 43.19 \\
    4B & 92.75 & 61.75 & 79.50 & 47.50 & 69.17 & 382.3 & 508.0 & 43.66 \\
    9B & 92.50 & 59.50 & 78.13 & 43.53 & 67.06 & 385.3 & 498.8 & 56.45 \\
    \midrule
    \multicolumn{9}{c}{\textit{JevAny-style (LoRA)}} \\
    \midrule
    0.8B & 92.50 & 71.50 & 78.63 & 47.06 & 70.39 & 230.8 & 305.2 & 17.49 \\
    4B & 92.25 & 69.50 & 80.13 & 44.12 & 69.65 & 298.1 & 410.3 & 22.15 \\
    9B & 92.00 & 60.50 & 80.63 & 42.65 & 67.76 & 292.3 & 394.3 & 25.55 \\
    \bottomrule
  \end{tabular}}
\end{table}

\paragraph{Router scale.}
\Cref{tab:router_scale} trains released StartLux checkpoints and JevAny-style routers without decision pretraining at 0.8B, 4B, and 9B on M5, all with the same data, reward, and LoRA budget; in neither family does a larger router bring a reliable gain, and the 0.8B router is the fastest.

\subsection{Ablation Study}
\label{sec:exp_ablation}

\begin{table}[t]
  \centering
  \caption{Ablations of \ours (StartLux; \%).
  Small: selection share of the small non-reasoning experts (2B/4B).}
  \label{tab:ablation}
  \setlength{\tabcolsep}{2.5pt}
  \resizebox{\linewidth}{!}{%
  \begin{tabular}{@{}lcccccc@{}}
    \toprule
    Variant & \shortstack{Math-\\Bench} & \shortstack{KOR-\\Bench} & \shortstack{OCR-\\Bench} & \shortstack{Follow-\\Bench-zh} & All & Small \\
    \midrule
    \ours (StartLux) & 96.20 & 58.40 & 83.00 & 46.41 & 70.22 & 32.80 \\
    (a) w/o quality-first adv. & 95.20 & 57.60 & 82.00 & 45.90 & 69.40 & 57.35 \\
    (b) w/o small-expert term & 95.80 & 57.20 & 82.44 & 47.05 & 69.93 & 5.26 \\
    (c) w/o reasoning-gain term & 95.80 & 58.00 & 82.89 & 46.79 & 70.15 & 24.85 \\
    (d) w/o visual input & 95.20 & 57.80 & 82.33 & 45.38 & 69.40 & 42.84 \\
    \bottomrule
  \end{tabular}}
\end{table}

\Cref{tab:ablation} removes one component at a time; every removal lowers All.
(a) Replacing the quality-first advantage with $A^{R}_i$ lets cost outweigh quality, as the quality-priority guarantee (\cref{sec:method_optim}) predicts: the small-expert share rises from 32.80\% to 57.35\% and all four benchmarks drop.
(b) Without the small-expert term, small experts are almost never selected (5.26\%), and (c) without the reasoning-gain term, All drops to 70.15.
(d) Blanking the router's images lowers all four benchmarks.

\section{Conclusion}
\label{sec:conclusion}

We introduced \ours, an adaptive router that learns a routing policy over cards that describe the current experts.
In the open-weight setting, it learns new tasks without replay where feedback-trained routers forget, puts added experts to use as its pool grows, learns a new pool from a small exploration budget, and keeps quality ahead of cost.
For billed API experts, it replaces RL with soft labels on feedback collected once, and the shared inputs and quality--cost trade-off remain effective under both losses.
On open-weight and API pools, it routes accurately at a fraction of the cost of single models.
A natural next step is adaptive routing when task distributions and expert pools change simultaneously, including expert retirement and repricing.

\section*{Limitations}

\ours routes each question to one frozen expert and does not cover multi-model collaboration or cascades.
Its adaptive updates learn from native feedback, so each new task requires an automatic scorer or judge, and each new pool a small exploration budget or, for API experts, one paid call per expert and training question.
We evaluate adaptation to new tasks and to an expanding pool as separate sequences; simultaneous task and pool changes, as well as expert retirement and repricing, which cards can express, are left to future work.
Our four-benchmark studies reuse questions analyzed during development, and our costs omit batching and end-to-end latency.

\bibliography{refs}

\clearpage
\appendix
\raggedbottom
\makeatletter
\setlength{\@dblfptop}{0pt}
\setlength{\@dblfpsep}{18pt}
\setlength{\@dblfpbot}{0pt plus 1fil}
\makeatother

\section{Router Designs and LoRA Variants}
\label{app:alternative_router}

The main tables report both routers with full finetuning.
Here we describe the JevAny-style router and compare full finetuning with LoRA for both designs.

\paragraph{Pointer-head parameterization.}
JevAny~\cite{jevany} encodes the request jointly with its candidates and scores each candidate with a pointer head.
Since the released JevAny model is 4B, we construct the 0.8B router from its published architecture on Qwen3.5-0.8B and initialize the query projection to zero, yielding a uniform initial policy with non-vanishing gradients.
It is trained without the decision pretraining of the released JevAny-4B.
Both router designs use the same reward and optimizer and are trained with full finetuning or LoRA for the main suite.

\subsection{Main-suite Results}

\Cref{tab:alternative_benchmark_results} compares full finetuning and LoRA on all 15 benchmarks under the protocol of \cref{tab:benchmark_results}.
LoRA lowers Avg. by 2.04 points for StartLux (63.04) and by 1.03 points for JevAny (62.01).
For StartLux it also cuts the cost from 12.56\% to 8.81\%, below every feedback-trained router in \cref{tab:benchmark_results}, while staying more accurate than all of them (at most 62.81); for JevAny, LoRA costs slightly more (12.14\% versus 10.88\%).

\begin{table*}[t]
  \centering
  \caption{Full finetuning versus LoRA on the main suite. Native scores and request-weighted Avg. (\%); Cost follows \cref{tab:benchmark_results}.}
  \label{tab:alternative_benchmark_results}
  \setlength{\tabcolsep}{3pt}
  \renewcommand{\arraystretch}{1.16}
  \resizebox{\linewidth}{!}{%
  \begin{tabular}{@{}l*{17}{c}@{}}
    \toprule
    & \multicolumn{5}{c}{\cellcolor[HTML]{FFF2CC}LLMRouterBench} & \multicolumn{6}{c}{\cellcolor[HTML]{DCEBF7}MMR-Bench V2} & \multicolumn{4}{c}{\cellcolor[HTML]{E4EFD8}RobustRouteBench} & \cellcolor[HTML]{FCE8E6} & \cellcolor[HTML]{F1E7FA} \\
     \multirow{-2}{*}{\raisebox{0.5ex}{Method}} & \cellcolor[HTML]{FFF2CC}\shortstack{Acc. $\uparrow$\\MathBench} & \cellcolor[HTML]{FFF2CC}\shortstack{Acc. $\uparrow$\\KOR-Bench} & \cellcolor[HTML]{FFF2CC}\shortstack{Acc. $\uparrow$\\SimpleQA} & \cellcolor[HTML]{FFF2CC}\shortstack{Acc. $\uparrow$\\FinQA} & \cellcolor[HTML]{FFF2CC}\shortstack{Acc. $\uparrow$\\MELD} & \cellcolor[HTML]{DCEBF7}\shortstack{Acc. $\uparrow$\\RealWorldQA} & \cellcolor[HTML]{DCEBF7}\shortstack{Acc. $\uparrow$\\OCRBench} & \cellcolor[HTML]{DCEBF7}\shortstack{Acc. $\uparrow$\\CharXiv} & \cellcolor[HTML]{DCEBF7}\shortstack{Acc. $\uparrow$\\MathVerse} & \cellcolor[HTML]{DCEBF7}\shortstack{Acc. $\uparrow$\\BLINK} & \cellcolor[HTML]{DCEBF7}\shortstack{aAcc. $\uparrow$\\HallusionBench} & \cellcolor[HTML]{E4EFD8}\shortstack{Acc. $\uparrow$\\HellaSwag} & \cellcolor[HTML]{E4EFD8}\shortstack{HSR $\uparrow$\\FollowBench-zh} & \cellcolor[HTML]{E4EFD8}\shortstack{Acc. $\uparrow$\\C-Eval} & \cellcolor[HTML]{E4EFD8}\shortstack{Acc. $\uparrow$\\AI2D} & \cellcolor[HTML]{FCE8E6}\multirow{-2}{*}{Avg.} & \cellcolor[HTML]{F1E7FA}\multirow{-2}{*}{Cost} \\
    \midrule
    \oursmethod{\ours (StartLux, Full FT)} & 96.20 & 58.40 & 5.80 & 37.60 & 60.20 & 69.24 & 83.00 & 57.70 & 72.67 & 61.20 & 72.55 & 89.20 & 46.41 & 81.00 & 82.00 & 65.08 & 12.56\% \\
    \oursmethod{\ours (StartLux, LoRA)} & 94.80 & 59.00 & 5.20 & 36.40 & 51.00 & 65.45 & 80.78 & 56.10 & 69.09 & 62.20 & 71.94 & 84.40 & 45.00 & 80.00 & 82.40 & 63.04 & 8.81\% \\
    \oursmethod{\ours (JevAny, Full FT)} & 95.20 & 58.20 & 4.20 & 36.00 & 55.80 & 67.67 & 81.33 & 54.70 & 66.92 & 61.20 & 71.14 & 85.80 & 44.49 & 80.20 & 82.20 & 63.04 & 10.88\% \\
    \oursmethod{\ours (JevAny, LoRA)} & 95.40 & 55.60 & 4.40 & 33.80 & 53.40 & 66.49 & 81.22 & 53.30 & 71.01 & 59.40 & 71.94 & 78.60 & 43.46 & 76.00 & 82.40 & 62.01 & 12.14\% \\
    \bottomrule
  \end{tabular}}
\end{table*}

\subsection{Task Adaptation}

\Cref{tab:alternative_continual_results} reports both training modes under the protocol of \cref{tab:continual_results}.
No variant loses more than 0.4 points on an earlier dataset relative to its best earlier stage.
With LoRA, the final five-dataset average is 82.3 for StartLux and 83.6 for JevAny, compared with 84.3 and 83.6 with full finetuning.

\begin{table*}[t]
  \caption{Task adaptation with full finetuning and LoRA (\%). Stage-4 parentheses: change from each dataset's first stage; Cost as in \cref{tab:continual_results}.}
  \label{tab:alternative_continual_results}
  \centering
  \begingroup
  \definecolor{stagegain}{HTML}{009C9C}
  \definecolor{stageloss}{HTML}{E53935}
  \definecolor{stageneutral}{gray}{0.45}
  \newcommand{\stagechange}[2]{\,{\scriptsize\textcolor{#1}{(#2)}}}
  \footnotesize
  \setlength{\tabcolsep}{2pt}
  \renewcommand{\arraystretch}{1.09}
  \resizebox{\textwidth}{!}{%
  \begin{tabular}{@{}l*{15}{r}@{}}
    \toprule
    & \multicolumn{2}{c}{\cellcolor[HTML]{FFF2CC}Stage 1} & \multicolumn{3}{c}{\cellcolor[HTML]{DCEBF7}Stage 2} & \multicolumn{4}{c}{\cellcolor[HTML]{E4EFD8}Stage 3} & \multicolumn{5}{c}{\cellcolor[HTML]{FCE8E6}Stage 4 (change vs.\ first)} & \multicolumn{1}{c}{\cellcolor[HTML]{F1E7FA}} \\
      \multirow{-2}{*}{\raisebox{0.5ex}{Method}} & \cellcolor[HTML]{FFF2CC}\shortstack{Acc. $\uparrow$\\MATH-\\500} & \cellcolor[HTML]{FFF2CC}\shortstack{Acc. $\uparrow$\\Chart\\QA} & \cellcolor[HTML]{DCEBF7}\shortstack{Acc. $\uparrow$\\MATH-\\500} & \cellcolor[HTML]{DCEBF7}\shortstack{Acc. $\uparrow$\\Chart\\QA} & \cellcolor[HTML]{DCEBF7}\shortstack{Acc. $\uparrow$\\MMLU-\\Pro} & \cellcolor[HTML]{E4EFD8}\shortstack{Acc. $\uparrow$\\MATH-\\500} & \cellcolor[HTML]{E4EFD8}\shortstack{Acc. $\uparrow$\\Chart\\QA} & \cellcolor[HTML]{E4EFD8}\shortstack{Acc. $\uparrow$\\MMLU-\\Pro} & \cellcolor[HTML]{E4EFD8}\shortstack{Acc. $\uparrow$\\Human\\Eval+} & \cellcolor[HTML]{FCE8E6}\shortstack{Acc. $\uparrow$\\MATH-\\500} & \cellcolor[HTML]{FCE8E6}\shortstack{Acc. $\uparrow$\\Chart\\QA} & \cellcolor[HTML]{FCE8E6}\shortstack{Acc. $\uparrow$\\MMLU-\\Pro} & \cellcolor[HTML]{FCE8E6}\shortstack{Acc. $\uparrow$\\Human\\Eval+} & \cellcolor[HTML]{FCE8E6}\shortstack{Acc. $\uparrow$\\IF\\Eval} & \multicolumn{1}{c}{\cellcolor[HTML]{F1E7FA}\multirow{-2}{*}{S4 cost}} \\
    \midrule
    \oursmethod{\ours (StartLux, Full FT)} & 92.20 & 85.84 & 91.80 & 85.44 & 74.00 & 92.00 & 85.52 & 73.80 & 84.10 & 92.20\stagechange{stageneutral}{+0.00} & 86.02\stagechange{stagegain}{+0.18} & 74.40\stagechange{stagegain}{+0.40} & 84.32\stagechange{stagegain}{+0.22} & 84.66 & 16.83\% \\
    \oursmethod{\ours (StartLux, LoRA)} & 84.60 & 83.38 & 89.40 & 84.59 & 71.80 & 90.60 & 85.52 & 73.40 & 78.66 & 90.80\stagechange{stagegain}{+6.20} & 85.68\stagechange{stagegain}{+2.30} & 73.00\stagechange{stagegain}{+1.20} & 78.66\stagechange{stageneutral}{+0.00} & 83.36 & 14.77\% \\
    \oursmethod{\ours (JevAny, Full FT)} & 83.20 & 83.58 & 90.20 & 84.91 & 71.80 & 91.00 & 84.63 & 72.80 & 80.49 & 92.00\stagechange{stagegain}{+8.80} & 85.84\stagechange{stagegain}{+2.26} & 74.40\stagechange{stagegain}{+2.60} & 81.71\stagechange{stagegain}{+1.22} & 83.92 & 16.73\% \\
    \oursmethod{\ours (JevAny, LoRA)} & 84.00 & 83.30 & 90.00 & 84.83 & 72.60 & 91.20 & 84.95 & 75.00 & 83.54 & 91.20\stagechange{stagegain}{+7.20} & 84.79\stagechange{stagegain}{+1.49} & 74.60\stagechange{stagegain}{+2.00} & 83.54\stagechange{stageneutral}{+0.00} & 83.92 & 15.62\% \\
    \bottomrule
  \end{tabular}}
  \endgroup
\end{table*}

\subsection{Trade-off across Router Designs}
\label{app:tradeoff_details}

Decreasing $\lambda$ from 0.30 to 0.01 raises JevAny-0.8B from 68.2\% to 74.8\% accuracy as its cost grows from 14.7\% to 64.7\% of Qwen3.5-9B (T) (\cref{fig:lambda}).
For $\lambda\ge0.10$, both routers stay within 68.2--70.4\% accuracy and 14.7--26.1\% cost; the gains at smaller $\lambda$ concentrate on FollowBench-zh and KOR-Bench, where JevAny-0.8B shifts its 9B share from the non-reasoning to the reasoning mode.

\section{API Router Training}
\label{app:api_training}

Both API routers (StartLux-0.8B and the JevAny-style router) are fully finetuned on the API training questions, with the ten datasets weighted equally.
As described in \cref{sec:method_api}, training reuses pre-collected expert feedback rather than repeatedly calling API experts during exploration; the loss below is \cref{eq:api_loss}, not the open-weight policy-gradient loss in \cref{eq:loss}.
With expert qualities $q_k$ and list-price costs $c_k$, the rewards are $u_k=q_k-\lambda c_k/\bar c$, where $\lambda=0.02$ and $\bar c$ is the mean training expert cost averaged over datasets; the target is $y=\operatorname{softmax}(u/\tau)$ with $\tau=0.05$, and with $p_k=\pi_\theta(e_k\mid x,\mathcal{E})$ the loss is
{\eqfontsize{9}
\begin{equation}
  \fitmath{%
  -\textstyle\sum_{k}y_k\log p_k+0.1\,\big(\max_{k}u_k-\sum_{k}p_k u_k\big),
  }
\end{equation}
}%
whose second term is the expected regret.
Training runs three epochs with 16 questions per update, AdamW (learning rate $10^{-6}$), and gradient clipping at 1.0; the candidate order is shuffled per question, and the router sees up to 16 distinct images.

\section{Hyperparameters}
\label{app:hyperparameters}

\Cref{tab:hyperparameters} lists the main and base four-benchmark training recipes; notation follows \cref{sec:method}. Cumulative pool-context replay uses the epoch-aligned budgets in \cref{app:pool_training_protocols}.
The main recipe trains StartLux-0.8B for \cref{tab:benchmark_results,tab:ablation} on about 100 questions per benchmark (1,700 in total), and each adaptation stage of \cref{tab:continual_results} applies it to that stage's new questions.
The fully finetuned JevAny-style router in \cref{tab:benchmark_results,tab:continual_results} uses the same reward and optimizer (\cref{app:alternative_router}).
The four-benchmark recipe trains a JevAny-0.8B pointer head on a frozen backbone for \cref{tab:pool,tab:data_scale}. In the sequential pool experiment, each stage adds 500 updates on the same 400 questions and inherits the preceding head and AdamW state, with a fresh linear learning-rate schedule. This recipe adds two training-only terms: a 4B (T) bonus that credits 4B (T) when it alone solves a question that all non-reasoning experts fail, and a KL penalty toward the uniform policy over candidates. M3 includes 9B (T) and its reasoning-gain bonus; only the 4B (T) bonus is absent. The exploration-budget experiment initializes its M10 routers independently. Both studies replay cached expert answers and scores instead of rerunning the experts, and each update uses only the rewards of the experts the router samples. The router-scale study (\cref{tab:router_scale}) applies the same questions and reward to M5, with rank-4 LoRA adapters (learning rate $5\times10^{-5}$) in place of the pointer head.
The full model in \cref{tab:ablation} was trained on one GPU and each ablation variant on eight.
ChartQA uses the revised scoring protocol; all other datasets use native scores.

\par\medskip\noindent\begin{minipage}{\linewidth}
  \centering
  \captionof{table}{Training hyperparameters.}
  \label{tab:hyperparameters}
  \small
  \setlength{\tabcolsep}{3pt}
  \resizebox{\linewidth}{!}{%
  \begin{tabular}{@{}lcc@{}}
    \toprule
    & Main & Four-benchmark \\
    \midrule
    Router & StartLux-0.8B & JevAny-0.8B \\
    Trainable parameters & all & pointer head \\
    Learning rate & $10^{-6}$ & $10^{-4}$, linear decay \\
    Training questions & 1,700 & 400 \\
    Epochs / updates & 1 / 107 & 20 / 500 \\
    Questions per update & 16 & 16 \\
    Attempts $G$ & 8 & 8 \\
    Exploration $\epsilon$ & 0.3 & 0.3 \\
    Cost weight $\lambda$ & 0.10 & 0.15 \\
    Small-expert bonus $\beta_s$ & 0.15 & 0.10 \\
    Reasoning bonus $\beta_t$ & 16 & 16 \\
    4B (T) unique-correct bonus & -- & 4 \\
    Advantage bound $\rho$ & 0.25 & 0.25 \\
    Entropy coefficient $\eta$ & 0.05 & 0.05 \\
    Uniform-KL coefficient & -- & 0.1 \\
    Seed & 37 & 37 \\
    \bottomrule
  \end{tabular}}
\end{minipage}

\section{Training Protocols for Expanding Pools}
\label{app:pool_training_protocols}

\Cref{tab:pool_training_protocols} compares three ways to train the same JevAny-0.8B router architecture: independent one-stage training for each pool, sequential training without previous-stage examples, and sequential training with cumulative replay.
The independent rows are the previously reported pool-size runs; the two sequential protocols share the same M3 checkpoint.
All use the same 400 distinct training questions and 2,280 test questions.
In cumulative replay, an example is a question paired with its original candidate pool: M5 mixes M3 and M5 examples, while M10 mixes M3, M5, and M10 examples.
Earlier examples retain their candidate cards and reward configuration; actions are sampled again using the current router.
The replay buffers contain 400, 800, and 1,200 question--pool examples, respectively, rather than additional distinct questions.
Training uses 20 epochs per stage, with 16 examples per update and eight action samples per example.
For cumulative replay, each epoch traverses the entire retained question--pool buffer: M3, M5, and M10 therefore require 500, 1,000, and 1,500 updates, respectively, and every retained question--pool example is visited exactly 20 times per stage.
Independent training and sequential training without replay use 500 updates per pool on 400 examples.
Thus, matching epochs gives cumulative replay more updates at M5 and M10.
Both sequential protocols inherit the pointer head and AdamW state and restart a linear learning-rate schedule over each stage's full training budget.
The epoch-aligned cumulative run starts M5 from the shared M3 checkpoint and M10 from its newly trained M5 checkpoint.
The final checkpoint is evaluated at each pool, without test-based checkpoint selection.
Accuracy increases from M3 to M10 under all three protocols.
Relative to sequential training without replay, cumulative replay raises All from 67.3\% to 70.0\% at M5 but lowers it from 72.1\% to 71.2\% at M10, despite the larger update budget.
These protocols illustrate that the router supports both one-stage training and sequential updates with or without cumulative replay; the main text retains the sequential results in \cref{tab:pool}.

\begin{table*}[t]
  \centering
  \caption{Training protocols for pool expansion. All: request-weighted.}
  \label{tab:pool_training_protocols}
  \small
  \setlength{\tabcolsep}{7pt}
  \begin{tabular*}{\textwidth}{@{\extracolsep{\fill}}l*{5}{c}@{}}
    \toprule
    Pool & \shortstack{Math-\\Bench} & \shortstack{KOR-\\Bench} & \shortstack{OCR-\\Bench} & \shortstack{Follow-\\Bench-zh} & All \\
    \midrule
    \multicolumn{6}{c}{\textbf{\textit{Independent (one-stage)}}} \\
    \midrule
    M3 & 93.5 & 64.0 & 78.4 & 38.7 & 66.7 \\
    M5 & 93.5 & 71.8 & 79.6 & 43.2 & 69.8 \\
    M10 & 93.8 & 72.3 & 85.9 & 54.6 & 75.5 \\
    \midrule
    \multicolumn{6}{c}{\textbf{\textit{Sequential}}} \\
    \midrule
    M3 & 93.0 & 67.0 & 78.9 & 38.4 & 67.2 \\
    M5 & 93.0 & 70.3 & 79.0 & 36.6 & 67.3 \\
    M10 & 92.3 & 74.8 & 84.5 & 44.3 & 72.1 \\
    \midrule
    \multicolumn{6}{c}{\textbf{\textit{Sequential + cumulative replay}}} \\
    \midrule
    M3 & 93.0 & 67.0 & 78.9 & 38.4 & 67.2 \\
    M5 & 92.8 & 74.5 & 80.4 & 41.9 & 70.0 \\
    M10 & 92.8 & 73.8 & 83.3 & 42.8 & 71.2 \\
    \bottomrule
  \end{tabular*}
\end{table*}

\section{Feedback Required to Add One Expert}
\label{app:incremental_feedback}

\paragraph{Protocol.}
Starting with one expert, we expand the active pool from one to five by adding one M5 expert at each of four stages.
The routers use a fixed five-card registry and mask unavailable experts before selecting an action.
We use five cyclic orders so that each expert enters at every pool size.
At each addition, learned routers start from the incumbent checkpoint and use a separate branch for each budget of 0, 10, 50, 100, 200, and 400 newly scored expert answers.
Budgets are nested and approximately balanced across the four benchmarks.
The count refers to judgments for the newly enabled expert. Incumbent feedback for those questions is already available.
The branch with 400 answers supplies the incumbent checkpoint for the next addition. Three seeds govern acquisition and training randomness.
At zero feedback, each trainable router skips the current update and routes within the incumbent pool using the incumbent checkpoint.
For the frozen baselines, we use the released StartLux 0.8B checkpoint and score five option tokens from a single multimodal prompt with Qwen3.5-0.8B.
We train the JevAny 0.8B starter on the 24 synthetic support tickets in the official starter recipe with its native multimodal adapter enabled, then freeze it throughout the expansion.
The frozen selectors choose their highest-scored available expert from the expanded pool with zero feedback.

\paragraph{Training and evaluation.}
Following the methods in \cref{tab:closed_source_results}, the learned comparisons use an MLP reward regressor, an adapted LatentRouter with train-only capability profiles, and one-decision PPO.
Both \ours arms use the quality-priority RLOO update of \cref{sec:method_optim} with a frozen backbone.
The JevAny variant uses a frozen Qwen3.5-0.8B backbone and trains its native pointer head from scratch. StartLux updates the five native letter readouts initialized from the released 0.8B checkpoint.
The three trained baselines use the frozen JevAny-0.8B representations. The PPO arm uses the JevAny pointer with a learned critic.
The common reward uses $2q-1$ and a penalty of $0.15$ per 4,096 generated tokens.
Each positive-budget branch warm-starts the previous parameters and uses a fresh AdamW optimizer for 40 batches of up to 32 questions, with two PPO updates per batch.
All routers select experts for the 2,280 test questions of \cref{tab:pool}. The paired offline comparison uses the 1,421 questions with archived native grades and output-token counts for every M5 expert (264 MathBench, 363 KOR-Bench, 351 OCRBench, and 443 FollowBench-zh).
The training feedback and test questions are disjoint.
We average results over five orders and, for trainable routers, three seeds.

\twocolumn[{%
\begin{minipage}{\textwidth}
  \centering\small
  \captionof{table}{Request-weighted accuracy (\%) by expansion and feedback budget. All averages the four expansions.}
  \label{tab:incremental_feedback_stages}
  \setlength{\tabcolsep}{3pt}
  \renewcommand{\arraystretch}{1.06}
  \resizebox{\textwidth}{!}{%
  \begin{tabular}{@{}lrrrrrrrrr@{}}
    \toprule
    & & & & \multicolumn{3}{c}{Trainable baselines} & \multicolumn{3}{c}{Frozen baselines} \\
    \cmidrule(lr){5-7}\cmidrule(l){8-10}
    Pool & Feedback & \shortstack{\ours\\(JevAny 0.8B)} & \shortstack{\ours\\(StartLux 0.8B)} & MLP & LatentRouter & Router-R1 & \shortstack{Prompted\\Qwen3.5-0.8B} & \shortstack{Original\\StartLux 0.8B} & \shortstack{JevAny 0.8B\\(Starter)} \\
    \midrule
    $1\to2$ & 0 & 67.3 & 67.3 & 67.3 & 67.3 & 67.3 & 71.5 & 66.9 & 67.8 \\
     & 10 & 67.0 & 69.1 & 67.6 & 65.8 & 66.3 &  &  &  \\
     & 50 & 69.7 & 71.2 & 68.8 & 66.7 & 68.9 &  &  &  \\
     & 100 & 71.4 & 71.9 & 69.7 & 67.6 & 67.4 &  &  &  \\
     & 200 & 71.5 & 72.2 & 70.3 & 67.0 & 70.7 &  &  &  \\
     & 400 & 72.2 & 72.1 & 69.9 & 66.3 & 70.4 &  &  &  \\
    \addlinespace
    $2\to3$ & 0 & 72.2 & 72.1 & 69.9 & 66.3 & 70.4 & 72.3 & 66.8 & 67.9 \\
     & 10 & 69.4 & 70.5 & 70.3 & 67.8 & 64.5 &  &  &  \\
     & 50 & 72.5 & 73.2 & 70.2 & 68.3 & 67.7 &  &  &  \\
     & 100 & 73.4 & 74.2 & 71.2 & 69.2 & 68.2 &  &  &  \\
     & 200 & 74.8 & 74.2 & 72.0 & 68.9 & 66.9 &  &  &  \\
     & 400 & 74.8 & 74.3 & 71.6 & 68.0 & 68.4 &  &  &  \\
    \addlinespace
    $3\to4$ & 0 & 74.8 & 74.3 & 71.6 & 68.0 & 68.4 & 73.0 & 66.6 & 67.8 \\
     & 10 & 69.3 & 71.9 & 70.2 & 68.3 & 66.9 &  &  &  \\
     & 50 & 72.2 & 73.7 & 69.8 & 67.9 & 67.7 &  &  &  \\
     & 100 & 73.8 & 74.0 & 70.8 & 68.3 & 68.6 &  &  &  \\
     & 200 & 73.8 & 74.7 & 71.0 & 67.9 & 69.4 &  &  &  \\
     & 400 & 74.5 & 74.8 & 71.6 & 68.8 & 69.4 &  &  &  \\
    \addlinespace
    $4\to5$ & 0 & 74.5 & 74.8 & 71.6 & 68.8 & 69.4 & 73.9 & 66.4 & 67.7 \\
     & 10 & 69.2 & 71.5 & 68.8 & 67.0 & 65.3 &  &  &  \\
     & 50 & 73.9 & 73.9 & 70.2 & 67.9 & 68.3 &  &  &  \\
     & 100 & 75.5 & 75.3 & 71.7 & 68.4 & 68.8 &  &  &  \\
     & 200 & 75.4 & 75.5 & 71.8 & 68.0 & 67.8 &  &  &  \\
     & 400 & 75.4 & 75.9 & 72.1 & 67.5 & 70.2 &  &  &  \\
    \midrule
    \textbf{All} & 0 & 72.2 & 72.1 & 70.1 & 67.6 & 68.9 & 72.7 & 66.7 & 67.8 \\
     & 10 & 68.7 & 70.8 & 69.2 & 67.2 & 65.8 &  &  &  \\
     & 50 & 72.1 & 73.0 & 69.8 & 67.7 & 68.2 &  &  &  \\
     & 100 & 73.5 & 73.8 & 70.8 & 68.4 & 68.3 &  &  &  \\
     & 200 & 73.9 & 74.1 & 71.3 & 68.0 & 68.7 &  &  &  \\
     & 400 & 74.2 & 74.3 & 71.3 & 67.6 & 69.6 &  &  &  \\
    \bottomrule
  \end{tabular}%
  }
\end{minipage}
\vspace{\dbltextfloatsep}%
}]

\paragraph{Sample threshold.}
A budget qualifies if it and all higher tested budgets score at most one accuracy point below the same router's 400-answer reference, use at most 10\% more output tokens, and score at most five points below its selection rate on questions only the new expert answers correctly.
We apply the last check where at least 20 such questions are available.
The threshold is the smallest tested budget that qualifies. A threshold of 400 means the reference was reached only at the last tested budget under this rule.
The threshold measures convergence to a router's own operating point.

\newpage

\paragraph{Results.}
Across the added experts, the upper median threshold ranges from 200 to 400 answers for JevAny and from 100 to 400 for StartLux.
At most 200 labels meet the criterion in 26 of 60 settings for JevAny and 32 of 60 for StartLux, compared with 24, 21, and 31 for MLP, LatentRouter, and PPO, respectively.
As each expert joins the pool, \ours updates the incumbent checkpoint (\cref{tab:incremental_feedback_stages}).
With 400 scored answers for the added expert, the update raises accuracy by 2.0 percentage points for JevAny and 2.2 for StartLux on average, with the largest gains at the first addition.

\twocolumn[{%
\begin{minipage}{\textwidth}
  \centering\small
  \captionof{table}{Routing results at the final $4\to5$ expansion. At zero feedback, trainable routers use M4 and frozen baselines use M5.}
  \label{tab:incremental_feedback_final_m5}
  \setlength{\tabcolsep}{3pt}
  \begin{tabular*}{\textwidth}{@{\extracolsep{\fill}}lrrrrrrr@{}}
    \toprule
    Router & Feedback & \shortstack{Math-\\Bench} & \shortstack{KOR-\\Bench} & \shortstack{OCR-\\Bench} & \shortstack{Follow-\\Bench-zh} & All & \shortstack{New-expert\\recall} \\
    \midrule
    \textbf{\ours (JevAny 0.8B)} & 0 & 95.7 & 76.8 & 81.1 & 54.9 & 74.5 & 0.0 \\
     & 10 & 94.9 & 66.9 & 80.4 & 46.9 & 69.2 & 27.1 \\
     & 50 & 95.3 & 75.8 & 80.4 & 54.2 & 73.9 & 24.6 \\
     & 100 & 95.6 & 76.5 & 81.0 & 58.2 & 75.5 & 21.1 \\
     & 200 & 96.1 & 77.4 & 80.9 & 56.9 & 75.4 & 16.7 \\
     & 400 & 95.6 & 78.1 & 80.7 & 57.0 & 75.4 & 18.6 \\
    \addlinespace
    \textbf{\ours (StartLux 0.8B)} & 0 & 95.8 & 74.3 & 82.2 & 56.9 & 74.8 & 0.0 \\
     & 10 & 95.5 & 70.9 & 80.8 & 50.5 & 71.5 & 39.3 \\
     & 50 & 96.1 & 73.5 & 81.8 & 54.6 & 73.9 & 18.1 \\
     & 100 & 96.1 & 75.2 & 82.1 & 57.5 & 75.3 & 25.3 \\
     & 200 & 95.7 & 75.6 & 82.6 & 57.7 & 75.5 & 17.1 \\
     & 400 & 95.8 & 76.1 & 82.1 & 58.9 & 75.9 & 23.3 \\
    \addlinespace
    MLP Reward Predictor & 0 & 94.9 & 71.0 & 80.6 & 51.2 & 71.6 & 0.0 \\
     & 10 & 95.4 & 64.9 & 80.3 & 47.0 & 68.8 & 31.3 \\
     & 50 & 95.1 & 67.5 & 80.1 & 49.8 & 70.2 & 29.4 \\
     & 100 & 94.6 & 69.8 & 80.6 & 52.4 & 71.7 & 27.2 \\
     & 200 & 94.6 & 70.5 & 80.2 & 52.7 & 71.8 & 28.3 \\
     & 400 & 95.0 & 71.5 & 80.6 & 52.3 & 72.1 & 24.5 \\
    \addlinespace
    LatentRouter (adapted) & 0 & 95.6 & 64.6 & 82.1 & 45.8 & 68.8 & 0.0 \\
     & 10 & 93.8 & 60.7 & 81.2 & 44.9 & 67.0 & 36.3 \\
     & 50 & 92.4 & 64.8 & 81.0 & 45.5 & 67.9 & 20.5 \\
     & 100 & 94.0 & 64.8 & 81.5 & 45.6 & 68.4 & 20.9 \\
     & 200 & 94.9 & 63.3 & 81.9 & 44.9 & 68.0 & 17.6 \\
     & 400 & 94.9 & 62.4 & 81.5 & 44.3 & 67.5 & 13.9 \\
    \addlinespace
    Router-R1 (PPO) & 0 & 94.4 & 66.1 & 81.2 & 47.8 & 69.4 & 0.0 \\
     & 10 & 94.3 & 61.2 & 80.2 & 39.7 & 65.3 & 24.6 \\
     & 50 & 94.8 & 63.9 & 81.4 & 45.8 & 68.3 & 13.8 \\
     & 100 & 94.7 & 64.3 & 81.2 & 47.3 & 68.8 & 14.0 \\
     & 200 & 94.3 & 62.7 & 81.5 & 45.3 & 67.8 & 12.4 \\
     & 400 & 94.6 & 66.9 & 81.3 & 49.7 & 70.2 & 13.5 \\
    \midrule
    Qwen3.5-0.8B (Prompted Selector) & 0 & 95.5 & 75.8 & 80.9 & 54.0 & 73.9 & 16.0 \\
    StartLux 0.8B (Original) & 0 & 92.0 & 65.0 & 78.1 & 42.9 & 66.4 & 22.4 \\
    JevAny 0.8B (Starter) & 0 & 93.2 & 62.5 & 78.6 & 48.1 & 67.7 & 30.9 \\
    \bottomrule
  \end{tabular*}
\end{minipage}
\vspace{\dbltextfloatsep}%
}]

At the final expansion (\cref{tab:incremental_feedback_final_m5}), adapting the four-expert checkpoint with 400 scored answers raises accuracy from 74.5\% to 75.4\% for JevAny and from 74.8\% to 75.9\% for StartLux.
We report request-weighted accuracy and new-expert recall as percentages. New-expert recall measures how often the router selects the added expert on questions only that expert answers correctly.

\end{document}